\documentclass[a4paper,10pt,twocolumn]{article}

\usepackage[margin=1in]{geometry}
\usepackage[T1]{fontenc}
\usepackage[utf8]{inputenc}

\usepackage{amssymb,amsmath,amsthm,amstext}

\usepackage{graphicx}
\usepackage{subcaption}
\usepackage{float}
\usepackage{placeins}
\usepackage[table]{xcolor}
\usepackage{booktabs}
\usepackage{makecell}

\usepackage[ruled,vlined,linesnumbered]{algorithm2e}

\usepackage[round]{natbib}
\usepackage{hyperref}

\usepackage[bottom]{footmisc}

\usepackage{orcidlink}

\newcommand{\Ferror}{$F_\textup{err}$}
\newcommand{\Berror}{$B_\textup{err}$}
\newcommand{\bruteforce}{\texttt{brute\_force}}

\newcommand{\topk}{\texttt{top\_k}}
\newcommand{\gridsize}{\texttt{cell\_size}}

\newcommand{\auc}[1]{\ensuremath{\mathrm{AUC}@#1^{\circ}}}

\newcommand{\realestateTenk}{RealEstate10k}
\newcommand{\eyeful}{EyeFul Tower}
\newcommand{\mvs}{MVS}
\newcommand{\probscore}{\texttt{probabilistic\_scoring}}
\newcommand{\globalpipe}{global}
\newcommand{\Globalpipe}{Global}
\newcommand{\incremental}{incremental}
\newcommand{\Incremental}{Incremental}
\newcommand{\auct}[1]{\makecell{AUC\\@#1$^{\circ}$}}

\newcommand{\principalPoint}{principal point}

\newcommand{\vggtptsOurs}{VGGT Pts + Our Opt}
\newcommand{\vggtff}{VGGT ff}
\newcommand{\rmaxx}{$R_\textup{max}$}

\newcommand{\pycolmap}{PyCOLMAP}
\newcommand{\regularsetting}{regular camera setting}

\newcommand{\REGULARSETTING}{Regular Camera Setting}
\newcommand{\fisheyesetting}{fisheye camera setting}

\newcommand{\FISHEYESETTING}{Fisheye Camera Setting}
\definecolor{verylightgray}{HTML}{E0E0E0}

\newcommand{\rowMarked}{\rowcolor{verylightgray}}

\let\oldcite\cite
\renewcommand{\cite}{\citep}
\newcommand{\citee}{\oldcite}

\newcommand{\subheading}{\subsubsection}

\usepackage{authblk}

\title{Robust Multi-view Camera Calibration from Dense Matches \footnotemark}

\author[1]{Johannes H\"agerlind\,\orcidlink{0009-0003-1150-3412}}
\author[1]{Bao-Long Tran\,\orcidlink{0000-0001-5028-7693}}
\author[1]{Urs Waldmann\,\orcidlink{0000-0002-1626-9253}}
\author[1]{Per-Erik Forss\'en\,\orcidlink{0000-0002-5698-5983}}

\affil[1]{Computer Vision Laboratory, Link\"oping University, Link\"oping, Sweden\\
\texttt{\{johannes.hagerlind, bao-long.tran, urs.waldmann, per-erik.forssen\}@liu.se}}

\date{}

\begin{document}

\twocolumn[
\maketitle
\vspace{-1.0em}
\abstract{
Estimating camera intrinsics and extrinsics is a fundamental problem in computer vision, and while advances in structure-from-motion (SfM) have improved accuracy and robustness, open challenges remain.
In this paper, we introduce a robust method for pose estimation and calibration.
We consider a set of rigid cameras, each observing the scene from a different perspective, which
is a typical camera setup in animal behavior studies and forensic analysis of surveillance footage.
Specifically, we analyse the individual components in a structure-from-motion (SfM) pipeline, and identify
design choices that improve accuracy.
Our main contributions are: 
(1) we investigate how to best subsample the predicted correspondences from a dense matcher to leverage them in the estimation process. 
(2) We investigate selection criteria for how to add the views incrementally.
In a rigorous quantitative evaluation, we show the effectiveness of our changes, especially for cameras with strong radial distortion ($79.9\%$ ours vs.\ $40.4\%$ vanilla VGGT).
Finally, we demonstrate our correspondence subsampling
in a global SfM setting where we initialize the poses using VGGT. The proposed pipeline generalizes across a wide range of camera setups, and could thus become a useful tool for animal behavior and forensic analysis.
}

\vspace{0.5em}
]

\footnotetext{This paper has been accepted for publication at the
21st International Conference on Computer Vision Theory and Applications (VISAPP 2026).
Conference website: \url{https://visapp.scitevents.org/}.}


\section{\uppercase{Introduction}}
\label{sec:intro}

A common setup in computer vision is to have a set of cameras in rigid configuration, with overlapping fields of view. Such a {\it view set} can be used to obtain accurate 3D measurements using photogrammetry.

\begin{figure}[!htpb]
    \centering
    \vspace{0.5em}

    \begin{subfigure}[b]{0.95\columnwidth}
        \includegraphics[width=\linewidth]{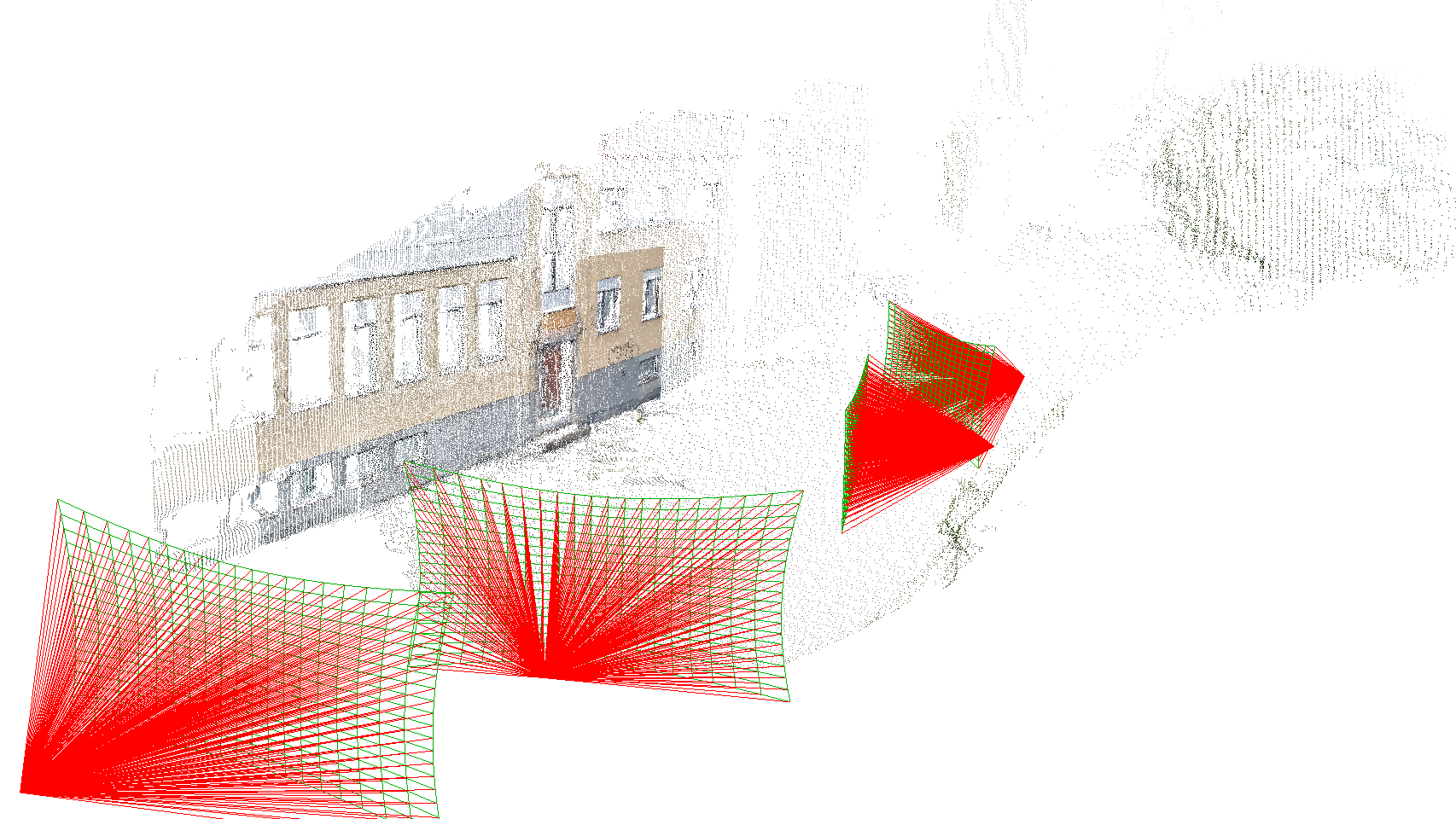}
    \end{subfigure}
    \begin{subfigure}[b]{0.24\columnwidth}
        \includegraphics[width=\linewidth]{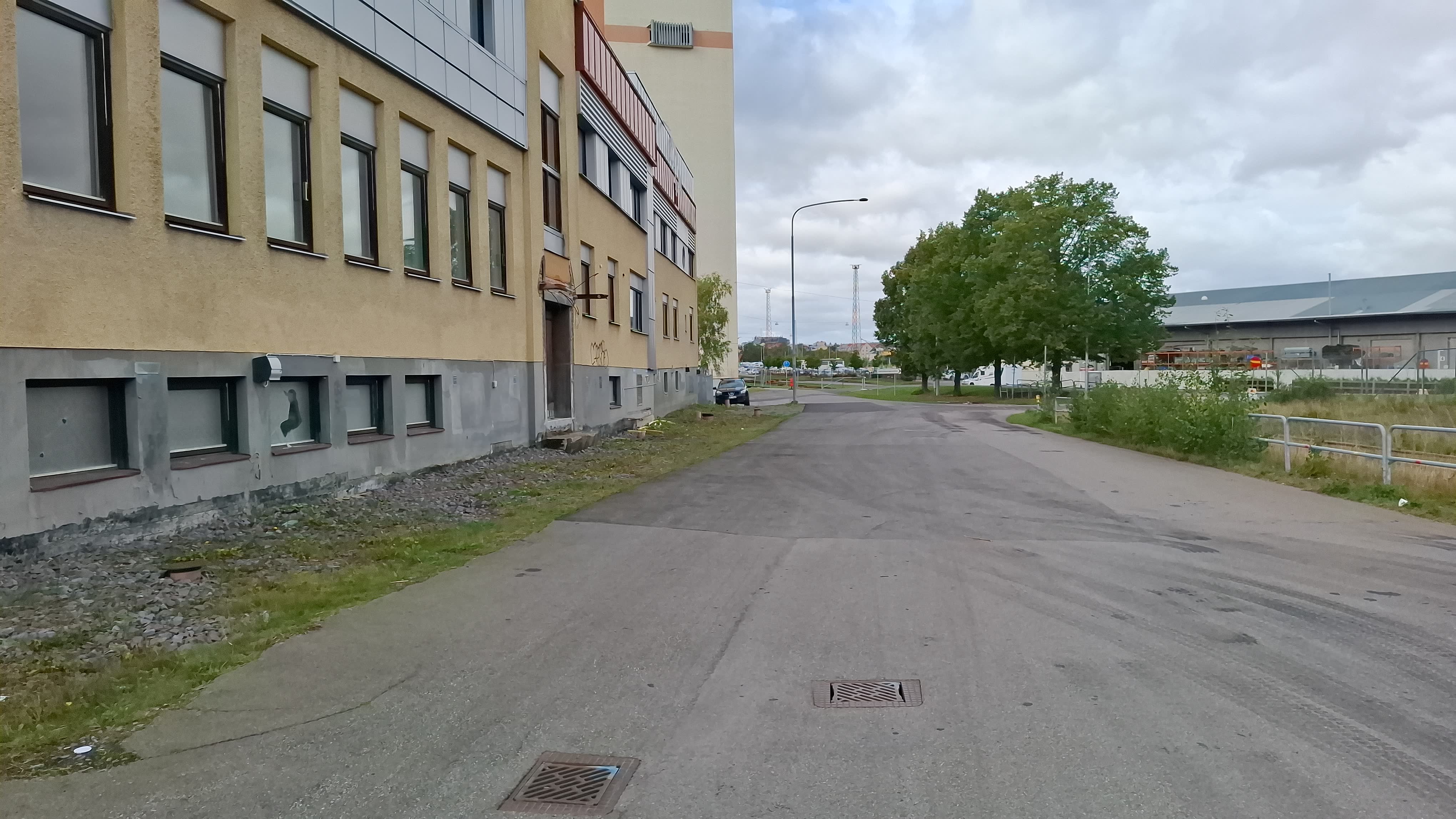}
    \end{subfigure}
    \begin{subfigure}[b]{0.24\columnwidth}
        \includegraphics[width=\linewidth]{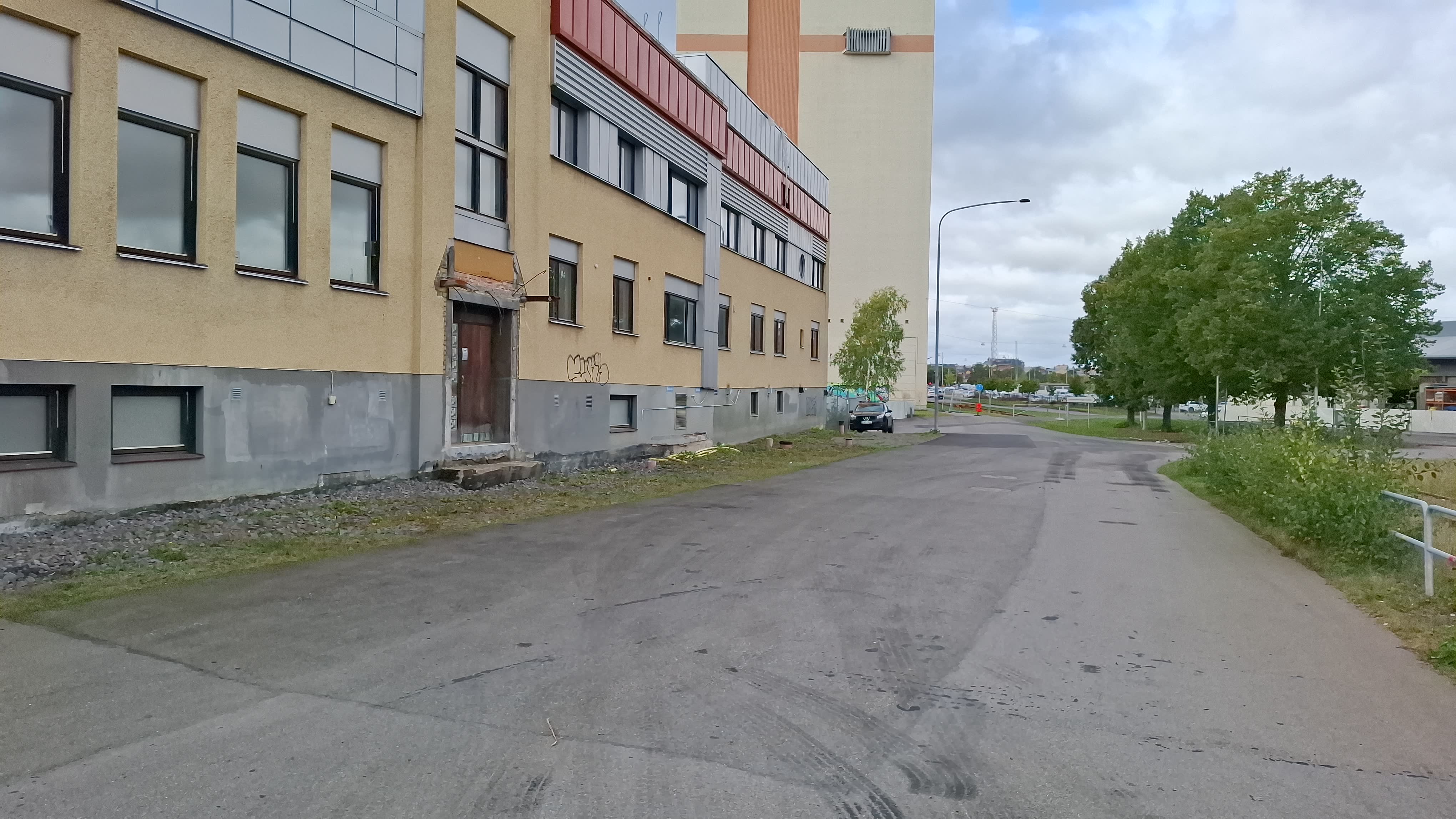}
    \end{subfigure}
    \begin{subfigure}[b]{0.24\columnwidth}
        \includegraphics[width=\linewidth]{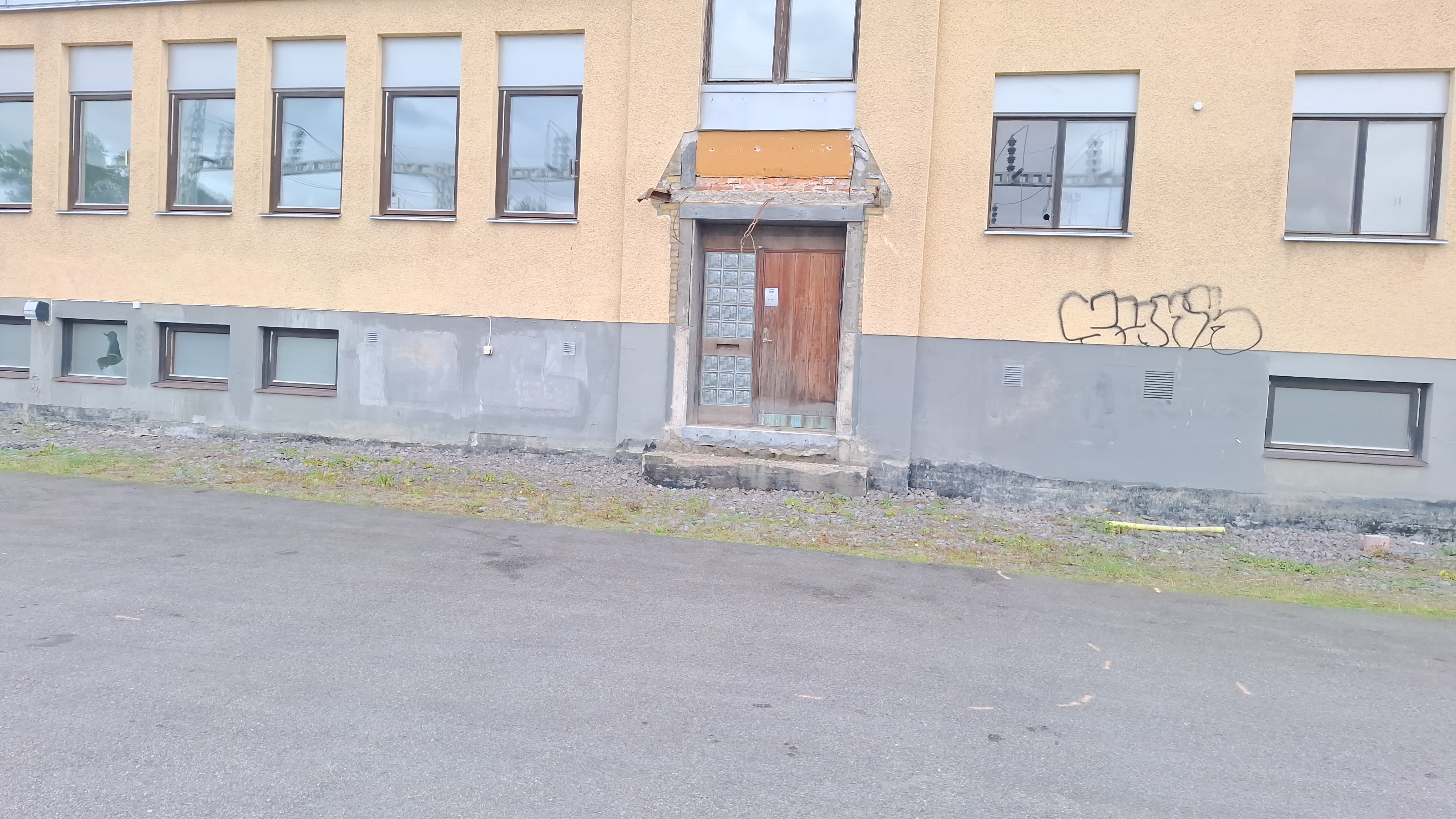}
    \end{subfigure}
    \begin{subfigure}[b]{0.24\columnwidth}
        \includegraphics[width=\linewidth]{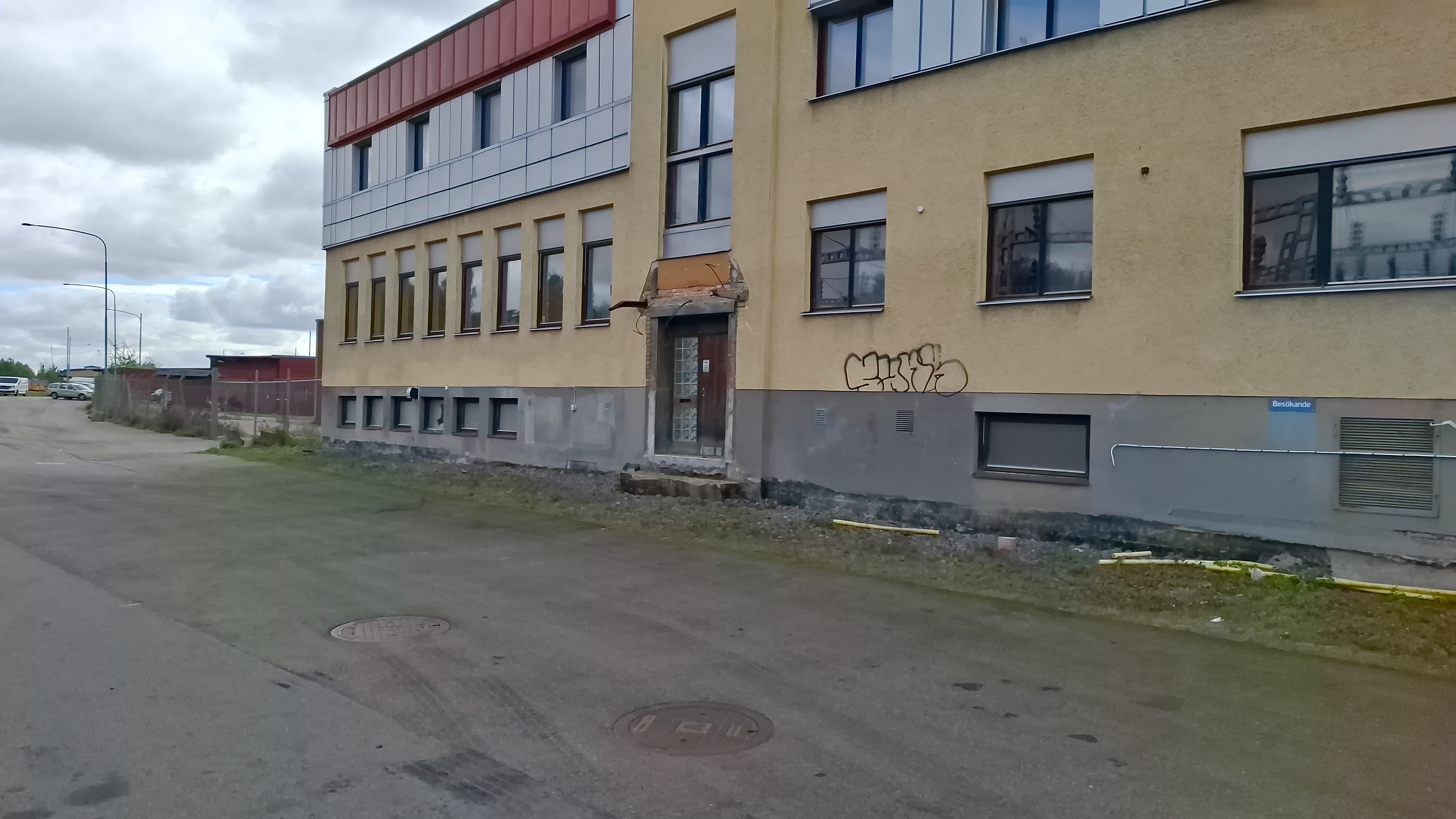}
    \end{subfigure}

    \caption{Sample forensic scenario where four mobile cameras capture a (simulated) crime scene. Top: Sparse reconstruction and camera calibration using our method. Bottom: Input images (photos by Henry Fr\"ocklin). }
    \label{fig:crime_scene}
\end{figure}

\begin{figure}[!htpb]
    \centering

    \vspace{0.5em}

    \begin{subfigure}[b]{0.95\columnwidth}
        \includegraphics[width=\linewidth]{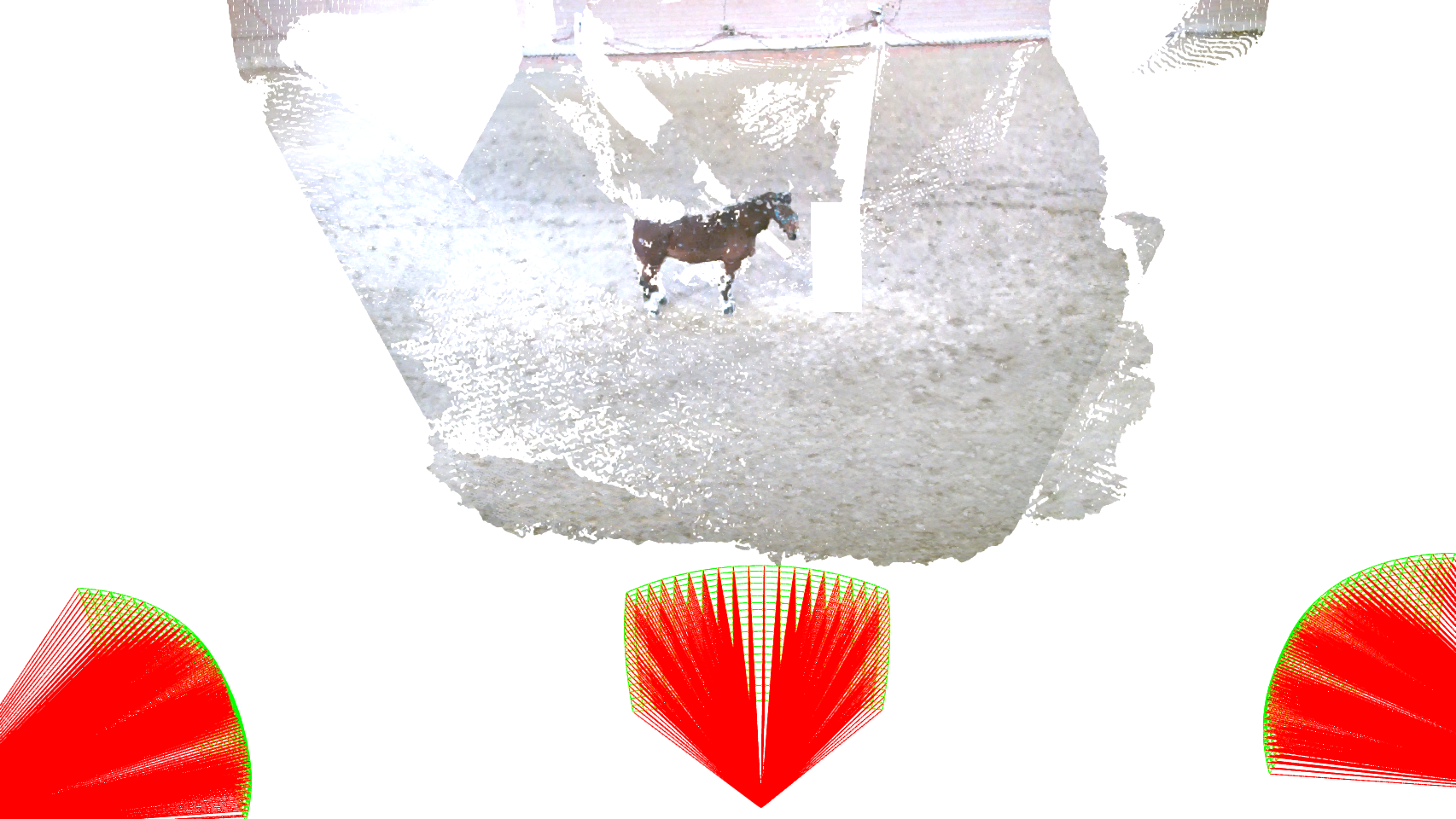}
    \end{subfigure}
    \begin{subfigure}[b]{0.31\columnwidth}
        \includegraphics[width=\linewidth]{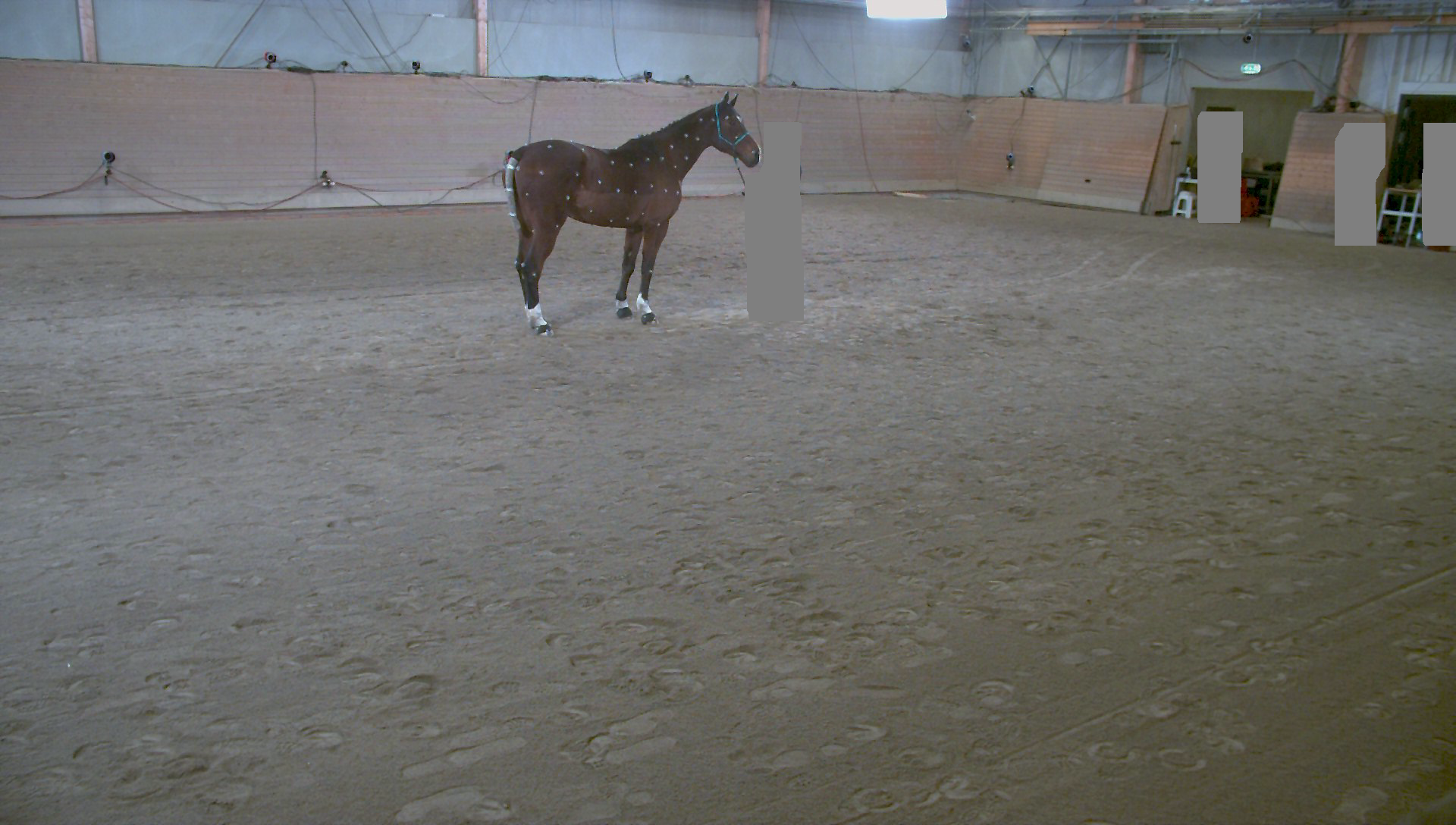}
    \end{subfigure}
    \begin{subfigure}[b]{0.31\columnwidth}
        \includegraphics[width=\linewidth]{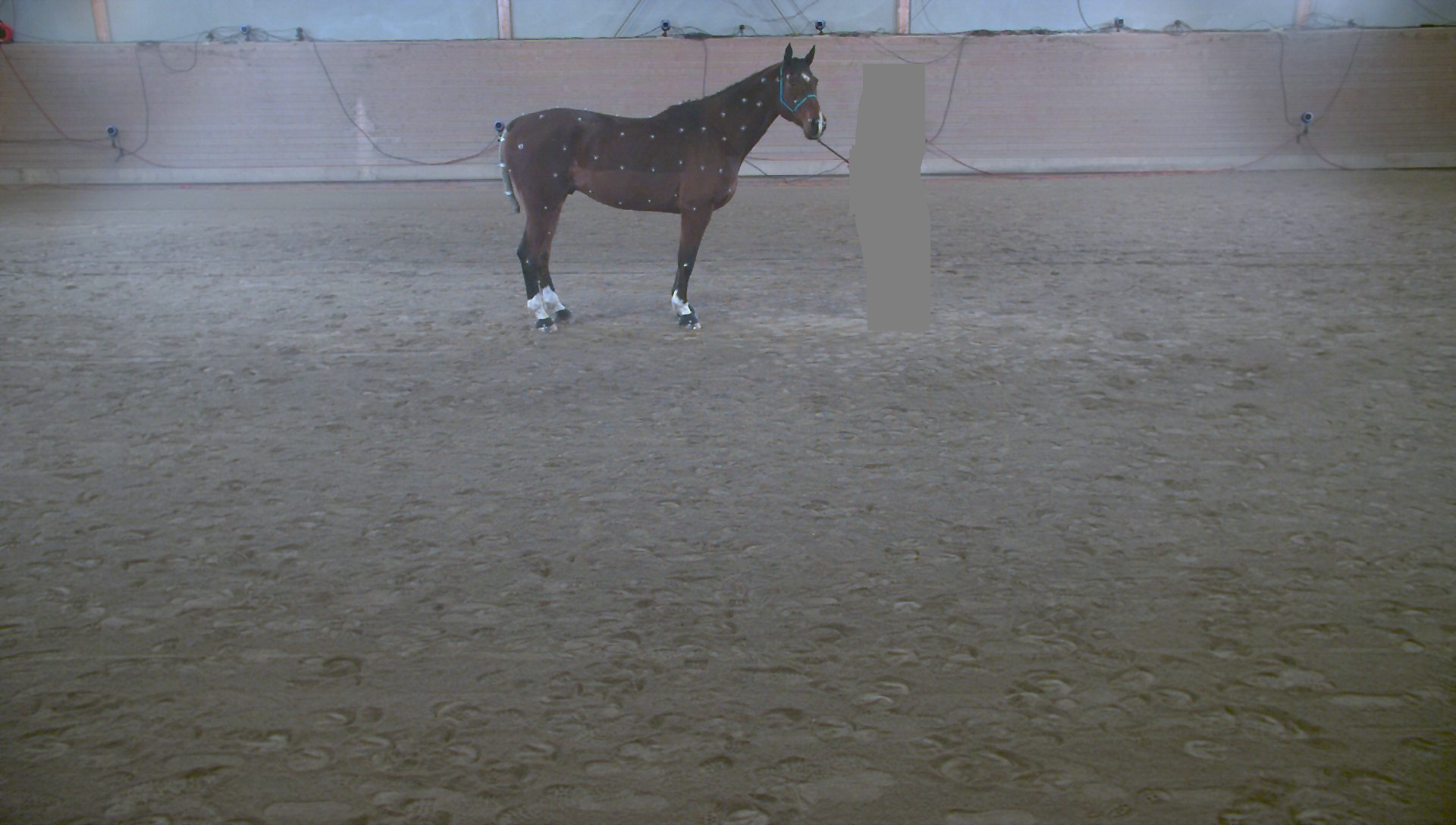}
    \end{subfigure}
    \begin{subfigure}[b]{0.31\columnwidth}
        \includegraphics[width=\linewidth]{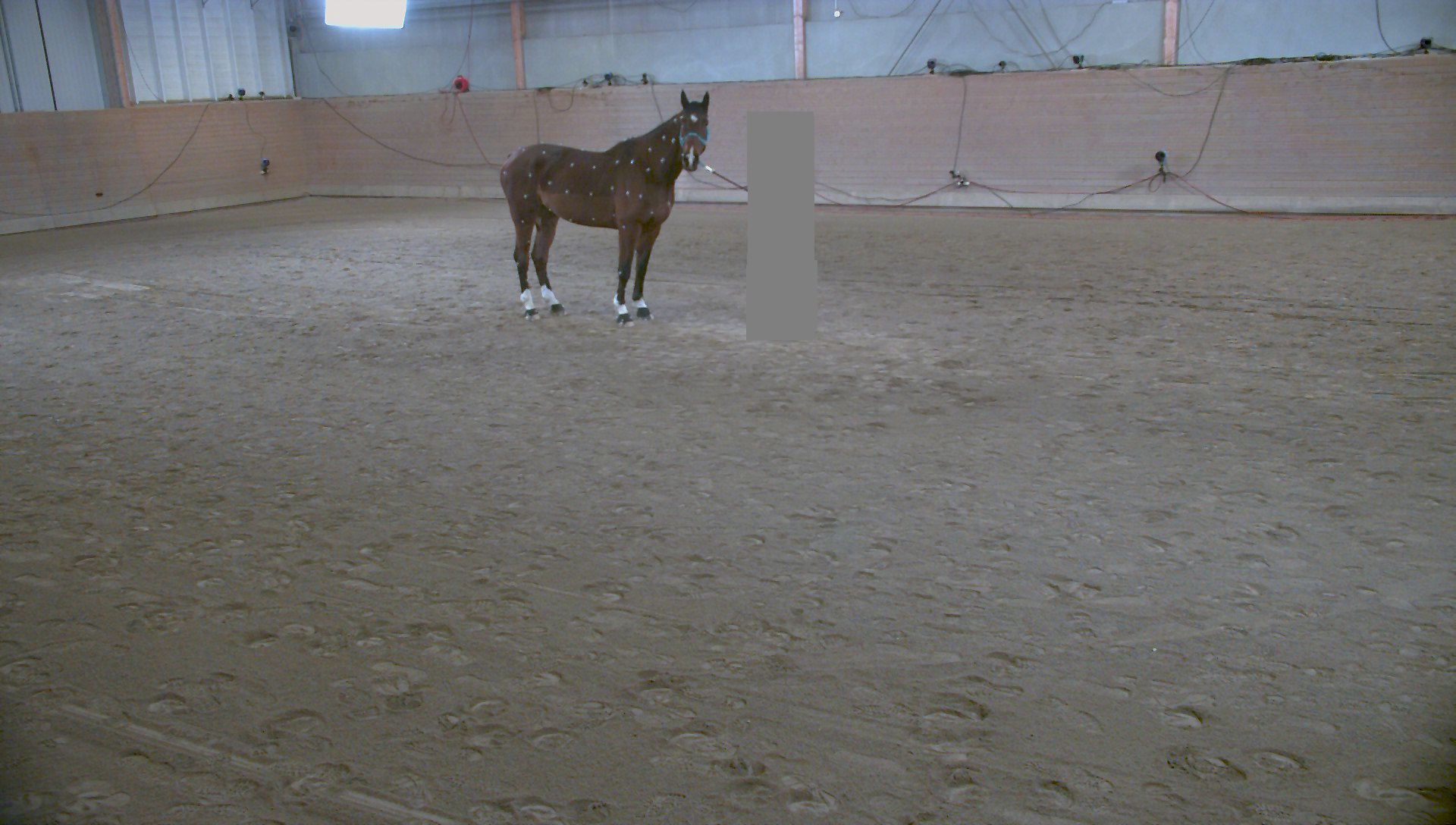}
    \end{subfigure}

    \caption{Sample animal behaviour scenario from the PFERD dataset \cite{li2024poses_pferd}.  Top: Point cloud reconstruction obtained by first applying the proposed method and then performing bundle adjustment over all matches while keeping the estimated camera intrinsics and extrinsics fixed.  Bottom: Input images. Note that the point-cloud contrast has been adjusted, and that people are removed for privacy reasons.}
    \label{fig:horse}
\end{figure}

A critical prerequisite for 3D sensing is that the view set is calibrated such that the pose and the full intrinsic calibration are known for each of the cameras. In many practical settings, however, the view sets are captured without the aid of a calibration pattern. 
They thus need to be {\it self-calibrated}, i.e.\ calibrated using natural landmarks in the scene.
%
In this paper, we explore the design space of 
multi-view self-calibration,
using dense correspondences in the images themselves, see Figs.~\ref{fig:crime_scene} and \ref{fig:horse} for examples.

The need for accurate camera poses and intrinsics frequently arise in field deployments, for instance with camera traps, where explicit scene calibration is rarely performed. Although an initial calibration can be performed, it may be rendered invalid by camera displacement, refocusing, or autofocus adjustments.
In animal behaviour analysis, accurate 3D information can be leveraged to estimate animals' skeletons~\cite{waldchan20243d} or to reconstruct their shape~\cite{SMALST}. 

Another application domain is forensic reconstruction, where surveillance or bystander footage from multiple viewpoints may need to be jointly analyzed to accurately localize individuals within a scene and thereby provide a more precise account of events~\cite{klasen21,Villa2019}  -- see Figures \ref{fig:crime_scene} and \ref{fig:horse} for example reconstructions.

Recent advances in structure-from-motion (SfM), such as VGGT~\cite{wang2025vggt}, have demonstrated state-of-the-art performance in camera pose estimation through feedforward architectures that directly predict camera poses, point maps, and depth maps. Nonetheless, these methods remain limited in their ability to handle cameras with significant lens distortion. Concurrently, progress in dense matching methods (e.g., RoMa~\cite{edstedt2024roma}, DKM~\cite{edstedt2023dkm}) motivates a re-examination of how traditional geometric approaches--based on triangulation, bundle adjustment, and related optimization techniques--compare to end-to-end neural models.

In this work, we specifically focus on the problem of camera pose estimation and self-calibration of view sets consisting of a few (3-10) views with wide relative baseline.
We explore the design space of algorithms, and investigate how classic geometry based methods and more recent deep learning methods can be combined to yield both robustness and accuracy.

\section{\uppercase{Related Work}}
\paragraph{Structure-from-Motion.}
Traditional incremental SfM methods, such as COLMAP~\cite{schoenberger2016sfm}, perform feature extraction, matching, and geometric verification. In two-view geometric verification, the inlier ratio between an estimated homography and a fundamental matrix is first evaluated. If this ratio falls below a threshold, the median triangulation angle is examined to assess whether the configuration is close to a pure rotation. Points from pairs identified as pure rotation are excluded from triangulation and thus not used for image registration.

Images are registered incrementally, one at a time, where each step includes triangulation, bundle adjustment and outlier removal. More recent approaches such as GLOMAP~\cite{pan2024glomap} do not use incremental registration. Instead, they first estimate the global camera orientations and positions from pairwise constraints, followed by a single global bundle adjustment.

\paragraph{Machine Learning Models.}
Recently, learning based alternatives to the traditional methods are starting to appear. These include DUSt3R~\cite{wang2024dust3r}, MASt3R~\cite{leroy2024mast3r}, VGGT~\cite{wang2025vggt} and Visual Geometry Grounded Deep Structure From Motion (VGGSfM)~\cite{wang2024vggsfm}.

Several papers have investigated how to use dense matches for SfM-related tasks. Most notably, \citee{astermark2025dense} present a dense match summarization method that is able to significantly reduce the runtime on 2-view relative pose estimation problems. 
In Dense-SfM~\cite{lee2025dense_sfm} the outputs of RoMa~\cite{edstedt2024roma} are cleaned up using bidirectional verification, i.e.\ a point in image A that maps to image B should also map back (close) to the point in image A. After this verification bundle adjustment is run. The 3D points are then projected to other views and a Gaussian splatting visibility filter checks if the 2 length track obtained from the bidirectional verification step can be extended to longer tracks. These tracks are then refined using a track-refinement network and finally bundle adjustment is run to obtain the final result.
\citee{martinec2007robust} develop a method that in the calibrated case can use as few as 4 points to represent a pairwise reconstruction. The proposed method also makes the assumption that the scene contains large surfaces where the clustering can be performed.

In our approach, we first run a match summarization scheme and then run SfM (either incremental or global) based on the summarization. This is similar to \citee{astermark2025dense} that first run a two-view summarization to speed up the more expensive estimation of two-view geometry. We investigate whether or not the matches from the dense matcher alone, after simple preprocessing and using some heuristics, can be used to get good camera pose estimation in a multi-view rig setup. Since in rigid multi-camera setups accurate calibration is often more important than speed, we focus on a robust method rather than a quick one.

\section{\uppercase{Method}}
In this section, we describe our system for pose estimation and calibration. We first describe dense correspondence sampling, then the structure of the incremental pipeline, including initial intrinsics and view order selection. We conclude with the global SfM setup and implementation details.

\subsection{Correspondence Cycle Sampling}
\label{sec:correspondance_cycle_sampling}
We use \emph{RoMa} \cite{edstedt2024roma} for dense matching. RoMa produces a dense, per-pixel warp-map along with an associated confidence score map. RoMA operates on a coarse resolution and outputs an upsampled resolution, which defines the resolution of the warp.  

We filter correspondences from the warp map by computing \emph{$n$-cyclic distances}, obtained by iteratively applying the warp until returning to the original image. A cycle is said to \emph{close} if, after $n$ applications of the warp, a point maps exactly back to its original coordinate. For two-cycles, we therefore retain only correspondences that return to the starting point after one forward and one backward mapping. For higher-order cycles $(n > 2)$, we discard any cycle that does not return to the starting coordinate, and additionally require that all of its $(n-1)$-subcycles also close. This strict criterion works well with RoMa as it preserves a large number of reliable matches. However, RoMa produces sub-pixel correspondences, which we handle by rounding coordinates to the nearest integer.

Using all correspondences in bundle adjustment is computationally prohibitive and tends to bias the reconstruction toward regions with high match density. To address this, we subsample the filtered correspondences to achieve a more balanced spatial distribution. Specifically, we overlay the warp output with a uniform grid of square cells of size $d \times d$. From each cell, we retain at most one correspondence. 

This design is motivated by \citee{schoenberger2016sfm}, who argues that a more uniform distribution of points improves stability when estimating unknown intrinsics. Also, \citee{astermark2025dense} showed that grid-based summarization achieves relative pose accuracy comparable to more complex clustering strategies, while being significantly faster.

\paragraph{Hierarchical Sampling.}
Alg.\ \ref{alg:cyclic_sampling} shows an overview of our hierarchical sampling. Sampling proceeds hierarchically by cycle order: grid cells are first filled with correspondences from $N$-cycles, then from $(N-1)$-cycles, and so on down to two-cycles if additional samples are required. Once a grid cell is filled by a higher-order cycle, no further points from that cell are considered. This prioritization favors longer tracks, which are particularly valuable for robust SfM reconstruction since it makes the bundle-adjustment system more tightly coupled \cite{triggs1999bundle}.

In our experiments, we consider cycles up to order four. This choice balances runtime efficiency with reconstruction quality: while checking only shorter cycles keeps the computation tractable, longer tracks can still emerge indirectly (e.g., if a point participates in both a 4-cycle and a 3-cycle, it may contribute to a track of length seven). However, such extended tracks become less common as grid size increases.

To ensure that views with matches concentrated in small regions are still represented, we require that at least $n_\textup{min} = c_{min} \cdot d^2$ points are ``filled'', where $d$ is the cell size (including cells filled by higher-order cycles). If this condition is not satisfied, we halve the grid cell size and repeat the procedure (but keep $n_\textup{min}$ from top iteration), up to a maximum of three iterations for a given cycle. We set the $c_{min}=0.3$ which 
results in
a good trade-off between ensuring coverage in views with locally concentrated matches and avoiding oversampling. 

\paragraph{Two Grid Levels.} We perform sub-sampling in two stages. In the first stage, we apply a fine grid of cell size $5 \times 5$ pixels. Within each cell, we randomly select a correspondence that satisfies the $n$-cycle condition. This step primarily reduces the number of matches and thereby lowers the computational cost of subsequent operations (e.g., two-view estimation and triangulation). 

In the second stage we input the cycles sampled from stage one and then we apply a coarser grid and select one correspondence per cell based on a scoring function (see Sec.\ \ref{sec:score_sampling}). Selection can be performed either deterministically, by choosing the correspondence with the highest score (argmax), or probabilistically, where each correspondence is sampled with probability proportional to its score normalized over all candidates in the cell. Only correspondences that satisfy the $n$-cycle condition are considered at this stage.

Assuming cycle orders and scores are precomputed, the worst-case complexity of the Stage~2 sampling in Algorithm~\ref{alg:cyclic_sampling} is
\(O\!\left(|S_1| \cdot \text{max\_iter} \cdot (N-1)\right)\),
where \(|S_1|\) is the number of correspondences retained after Stage~1 (worst case \(|S_1| = O(W H /g_{\text{fine}}^2)\), with \((W, H)\) the RoMa output resolution and \(g_{\text{fine}}\) the fine-grid cell size), \(\text{max\_iter}\) is the number of grid refinements, and \(N\) the maximum cycle order. Stage~2 scans all candidates per coarse cell (e.g., via \(\arg\max\) or score-proportional sampling). 
Runtime is linear in the number of images.

\begin{algorithm}[ht!]
\small
\caption{Hierarchical cycle sampling with two-stage grid subsampling.}
\label{alg:cyclic_sampling}
{
\ttfamily
\KwIn{Correspondences $C$ with cycle order, max cycle $N$, initial grid $d$, $c_\text{min}$, max iterations}
\KwOut{Selected correspondences $S$}
$S \gets \emptyset$\;

\For{$n = N$ down to 2}{  
    $d_\text{curr} \gets d$\;
    \For{iter = 1 to max\_iter}{  
        define cells with size $d_\text{curr} \times d_\text{curr}$\;
        \For{each cell $g$ in grid with $\text{candidates} \in \text{select\_order(n,C)}$ }{
            \If{$g$ not filled by higher-order cycle}{
                Stage 1 (fine grid): select one correspondence randomly\;
                Stage 2 (coarse grid): select by argmax(score) or probabilistic $\propto$ score\;
                add selected correspondence to $S$\;
            }
        }
        \If{number of filled cells $\ge n_\text{min} = c_\text{min} \cdot d^2$}{
            break \tcp{coverage sufficient}
        }
        \Else{
            $d_\text{curr} \gets d_\text{curr} / 2$ \tcp{refine grid}
        }
    }
}
}
\Return $S$\;
\end{algorithm}

\subsection{Score Sampling}
\label{sec:score_sampling}
\emph{Bundler} by \citee{bundler_IJCV08},  performs triangulation for all image pairs and ensures that at least one pair within a track has a sufficient triangulation angle. 
In contrast, \citee{schoenberger2016sfm} propose a multi-view triangulation strategy based on RANSAC, demonstrating both increased efficiency and improved reconstruction accuracy. 

In our setting, the objective is to select the most reliable correspondences from among multiple candidates that are likely to be correct. 
This task becomes particularly challenging when the image sequence is dominated by camera rotations resulting in small triangulation angles. 
In such cases, initialization methods such as VGGT can still produce useful results despite the narrow baseline, making it undesirable to impose a fixed angular threshold. 
To overcome this limitation, we introduce a novel scoring scheme. 

Our approach is motivated by practical heuristics from the NASA Ames Stereo Pipeline \cite{asp_guidelines_2025,beyer2018ames}, which suggest that triangulation angles between $10^\circ$ and $60^\circ$ provide reliable reconstructions in practice. 
Based on this observation, we define a \textbf{score function} for triangulation angles. 
The function attains its maximum at $\alpha=30^\circ$ and is modeled as a \textbf{symmetric Gaussian} density $G(\theta-\alpha,\sigma)$, with a standard deviation of $\sigma=20^\circ$, thereby approximately covering the interval $10^\circ$-$50^\circ$. 
Each pairwise correspondence is assigned a score by evaluating the Gaussian density function at its estimated triangulation angle, $\theta$. 

\begin{equation}
\begin{aligned}
f(\theta; \alpha, \sigma, p) 
&= \frac{p}{\displaystyle \max_{\theta \in [0,\pi]} \bigl\{ B(\theta) \, G(\theta) \bigr\}} \; B(\theta)\, G(\theta), \label{eq:score_function} \\
\text{where} \quad B(\theta) &= \exp \!\left(-\frac{1}{\tfrac{\theta}{\pi}\bigl(1 - \tfrac{\theta}{\pi}\bigr)}\right), \;\text{and}\\
G(\theta) &= \exp\!\left(-\tfrac{1}{2}\,\left(\tfrac{\theta - \alpha}{\sigma}\right)^2\right).
\end{aligned}
\end{equation}
Here $B(\theta)$ enforces boundary behavior such that the score vanishes as $\theta \to 0$ or $\theta \to \pi$, and $p$ is a normalization factor ensuring that $\max_\theta f(\theta) = p$.

The score function \eqref{eq:score_function} is evaluated for each pair within an $n$-cycle. 
The final score of a correspondence is obtained by summing the top-$k$ scores among all pairs in the $n$-cycle containing that correspondence. 
We evaluate settings with $k \in \{1, 2, 3\}$, as well as $k = 0$, where no scoring is applied and pairs are instead sampled uniformly within the cycle. 

\subsection{Incremental SfM}

\subheading{Initial Focal Length}
\label{sec:init_focal}
To determine the focal length we use a similar method as \citee{self-calibration}. \citee{huang2002some} proved that 2 singular values of the essential matrix has to be same and that the last one has to be 0. 
The residuals we use are thus defined as:

\begin{equation}
r_{ij} = w_{ij} \left(\frac{\sigma_{ij}^{(1)} - \sigma_{ij}^{(2)}}{\sigma_{ij}^{(1)} + \sigma_{ij}^{(2)}} + \sigma_{ij}^{(3)}\right)
\end{equation}
where $w_{ij}$ is a weight, and $\sigma_{ij}^{(1)}\geq  \sigma_{ij}^{(2)}\geq \sigma_{ij}^{(3)}$ are the ordered singular values of $E_{ij} = K_j^T F_{ij} K_i$, where $F_{ij}$ is the fundamental matrix. 

We can then optimize the focal lengths of $K_j$ and $K_i$ with respect to this cost function. 
For efficiency we use non-linear least squares on the $M^2$ element residual vector (where $M$ is the number of views) instead of summing them as in \cite{self-calibration}.

Other differences from the original paper is that we sum $\sigma_{ij}^{(1)}$ and $\sigma_{ij}^{(2)}$ in the denominator and add $\sigma_{ij}^{(3)}$. By adding $\sigma_{ij}^{(3)}$ we encourage the smallest singular value to go to zero. For more details see \cite{self-calibration}. We chose to set the focal length $f_x=f_y$ \cite{wang2023posediffusion, ding2024fundamental} for each camera (but they can vary between cameras). We set $w_{ij}$ to be the fraction of the number of matches used to calculate a specific $F_{ij}$ matrix and the maximum number of matches for all  $M$ view pairs. 

The $F$ matrices that are needed for the optimization are calculated using OpenCV implementation of MAGSAC++ \cite{barath2020magsac, opencv_library}.

\subheading{View Order}
To determine the view order, we calculate the score of each three-cycle (see Sec.\ \ref{sec:score_sampling}). We begin by selecting the three nodes that form the highest-scoring cycle. After that, we iteratively add the node that yields the greatest increase in total score.

For example, suppose we already have four views in the sequence and we want to add a new node. In this case, four possible three-cycles can be formed that include the new node. We compute the sum of their scores and compare this to the scores obtained by adding any other candidate node. The node that maximizes the score is selected.

This strategy resembles choosing the next view that shares the most points with the existing structure-from-motion (SfM) problem, but it differs in that we also apply a weighting based on our scoring function.

It was observed that the ordering of the initial three views can significantly
affect the reconstruction. While fixing some points early is known to reduce
the risk of bad local optima in incremental SfM, in practice we observed that concave or convex
scenes still caused failures. To address this, we apply a \textit{brute-force}
strategy: evaluate all three possible ways to select the first pair and then select the one with the lowest cost according to:

\begin{equation}
    (i^*, j^*) = 
    \textup{argmin}_{(i,j) \in \{(1,2),(1,3),(2,3)\}}
    \frac{1}{N_{ij}^2} \sum_{k=1}^{N_{ij}} d_{ij}^{(k)}
\end{equation}
where $N_{ij}$ is the number of points in front of both cameras for pair $(i,j)$,
and $d_{ij}^{(k)}$ is the reprojection distance of the $k$-th point. This is essentially the mean reprojection distance divided by the number of points, to put more weight on pairs that has many points on the correct side of the camera.

\subsection{Global SfM}
\label{sec:method_global_sfm}
To stabilize the bundle adjustment we fix the extrinsics of the camera with the highest sum of scores according to Sec.\ \ref{sec:score_sampling}. When summing the scores we only consider points that participate in a cycle of at least length 3.

We use the camera poses from VGGT as initialization.
Then we use match summarization obtained in the same way as in the incremental pipeline and initialize the 3D points from multiple views using robust triangulation \cite{schoenberger2016sfm}. This is followed by three bundle adjustment steps where in the first step (100 iterations) we only refine extrinsics, in the second step (200 iterations) we add refinement of focal length and in the final step we add the extra parameters (e.g. radial distortion etcetera).

\section{\uppercase{Experiments}}
We conduct extensive experiments to analyze the effect of different hyperparameter settings and design choices in our two SfM pipelines. In particular, we present ablation studies across multiple datasets with varying scene characteristics. Two of the datasets contain images captured with a camera with minimal distortion - we call this setting \textit{regular setting}. To see how our pipelines work in cases with strong regular distortion we use a dataset that contains images captured using fisheye cameras. We call this setting \textit{fisheye setting}.

\subsection{Metrics}
Following \citee{wang2023posediffusion,wang2025vggt}, we evaluate the performance of our proposed method using the AUC metric, which is calculated from Relative Rotation Accuracy (RRA), and Relative Translation Accuracy (RTA) metrics. Both RTA and RRA are angular errors, and the AUC@\texttt{threshold} metric jointly reflects both RRA and RTA under the same threshold (in degrees).

\subsection{Implementation Details}
\label{sec:implementation_details}
We use \texttt{\pycolmap{}
} 3.11.1 \cite{schoenberger2016sfm} for both the \incremental{} and the \globalpipe{} pipeline. In both pipelines Bundle adjustment (BA) is performed with a Cauchy loss using default settings (see \citee{triggs1999bundle} for details). We allow a relatively large number of iterations to reduce the risk of stopping BA before convergence is reached. In the global pipeline we allocate additional iterations, as we initially observed slower convergence in this setting. Note, however, that BA might terminate earlier if the convergence criteria are satisfied.

Unless otherwise stated, we use a pinhole model with $f_x = f_y$ and two radial distortion parameters. In the \fisheyesetting{} (e.g., the \textit{\eyeful{}} dataset - see Sec.~\ref{sec:experiments:fisheye_settings}), we use the FISHEYE\_RADIAL model in COLMAP and additionally attempt further refinement to the OPENCV\_FISHEYE model.\footnote{See COLMAP documentation for details.} To simulate a scenario with multiple distinct cameras observing the scene, we do not assume shared intrinsics, although for the two datasets in the \regularsetting{} (see Sec. \ref{sec:experiments:pinhole_cameras}) each individual scene is originally captured using the same physical camera.

The F matrices calculated using MAGSAC++ \cite{barath2020magsac} (to estimate the initial intrinsics, see \ref{sec:init_focal}) are used to filter the outliers. For each pair of point correspondence, we calculate the geometric distance from the point in image A to the epipolar line generated by the point in image B. We do this in both directions. We denote the threshold for this outlier removal as \textit{\Ferror{}}. In the regular camera setting, 
we set the \Ferror{} to 10 pixels, rather high, but robustification starts at 1 pixel (being the default setting in COLMAP 3.11.1), and in the \fisheyesetting{} we experiment with different settings of \Ferror{} (10, 100, 250, 500 and $\infty$).

When adding a new view in the \incremental{} pipeline, we use only previously registered points that form at least a three-cycle with the candidate view---that is, points that have already been triangulated and are observed by at least two registered views. The absolute pose is initialized using COLMAP's \texttt{estimate\_and\_refine\_absolute\_pose} function and refined over 100 iterations while optimizing focal length and radial distortion.

We then run bundle adjustment (BA) for 100 iterations using the added point. After that we include all points that form two-cycles with the new view and run another round of BA (again for 100 iterations). In both of these BA steps, we optimize the camera extrinsics, focal length, and radial distortion, but we do not refine the \principalPoint{}. After these adjustments, we remove all points with a pixel reprojection error greater than \textit{\Berror{}} = 20, following established best practices~
\cite{bundle_adjustment_asp, beyer2018ames}.

Once all views have been registered, we perform a final global BA (100 iterations), during which the principal point is also included in the refinement. While~\citee{schoenberger2016sfm} advises against refining the camera center for uncalibrated images, however, the FAQ of the documentation \cite{Schoenberger2025COLMAPFAQ} notes that doing so may be beneficial as a final global step---particularly when multiple cameras share the same intrinsic parameters.

In the \globalpipe{} pipeline, we start by refining the extrinsics during BA (300 iterations) - we do this since the initial estimation of the cameras' extrinsics can be wrong. We then add the focal length for 200 iterations and subsequently add the distortion parameters for another 200 iterations. This more fine-grained strategy for introducing parameters is used because the \globalpipe{} pipeline starts with all cameras and points already present in the scene. Like in the incremental pipeline, we run a final BA where the principal point is refined (100 iterations). In the \fisheyesetting{}, we attempt a final optimization where we initialize from the tracks, 3D points, and FISHEYE\_RADIAL camera model and refine it to a FISHEYE\_OPENCV camera model by refining the distortion parameters only for 100 iterations. In the \globalpipe{} pipeline, we do not impose a \Berror{} threshold, since all refinement steps are executed only once and are already backed up by a robust loss function. In contrast, the incremental pipeline benefits from an explicit threshold to ensure stability.

\paragraph{General Setting.}
 In each of the experiments we sample 10K points in each view set. We sample 6K four-cycles, 3K three-cycles and 1K two-cycles. Within each n-cycle (n=4,3,2), the distribution we sample depends on the scoring setting used as well as the distribution of pre-processed cycles (see Sec. \ref{sec:correspondance_cycle_sampling} and Sec. \ref{sec:score_sampling}).

\subsection{\REGULARSETTING{}}
\label{sec:experiments:pinhole_cameras}

\paragraph{Datasets.}
In the \regularsetting{} we evaluate two datasets; the \mvs{} dataset and the \realestateTenk{} dataset.
The {\it \mvs{} dataset} contains 124 multi-view scenes captured in a controlled environment, whereas {\it \realestateTenk{}} consists of frames sampled from handheld video sequences. For our experiments, we randomly select 124 scenes from the test split of \realestateTenk{}. We restrict our evaluation to this subset in order to explore a wide range of settings within a feasible computational budget. Throughout the remainder of the text, references to \realestateTenk{} denote this subset. 
To simulate our few-view scenario, we subsample the scenes by selecting 10 images per scene from each dataset, following the approach of \citee{wang2025vggt}.

A notable distinction between the two datasets is that \realestateTenk{} often contains scenes dominated by forward/backward translational motion and scenes dominated by rotational motion---both of which are known to present significant challenges for traditional SfM methods.  

\paragraph{Results.}

We evaluate both the incremental and global variants of our pipeline, cf. Tab.~\ref{tab:comparison:global_results_dtu_mvs}. 
For the MVS dataset we achieve an $\auc{30}$ of $99.4$ in the global pipeline and an $\auc{30}$ of $96.2$ in the incremental pipeline. 
On the \realestateTenk\ dataset our best setting achieves $\auc{30} = 88.4 $ in the global pipeline and $\auc{30}=67.4$ in the incremental one.

Comparing our \globalpipe{} pipeline with the VGGT feed-forward (\vggtff{}) method on the MVS dataset, we achieve the same \auc{30} score of 99.4 and a comparable \auc{3} score (Ours: 94.1 vs. VGGT-ff: 94.2). These results are obtained when we only refine the extrinsics in our \globalpipe{} pipeline. If we additionally refine the distortion parameters of the RADIAL camera model, we achieve a slightly higher \auc{3} but a slightly lower \auc{30}; see Sec. \ref{sec:ablation_pinhole} for details.
We also compare against the VGGT pipeline~\cite{wang2025vggt}, using their initialization (tracks and points) and then using either their (\vggtff{} + BA) or our optimization steps (\vggtptsOurs{}). Note that in both cases, both the \auc{30} and the \auc{3} is lower than for our \globalpipe{} pipeline. The drop of \auc{3} is very significant with -27.1 AUC scores if we compare our \globalpipe{} pipeline with \vggtff{} + BA (Ours: 94.1 vs. VGGT-ff + BA: 67.0). This suggests that our way of filtering and selecting points is important for the stability of the post-VGGT optimization (i.e.\ use our point summarization rather than their point tracks and point initilization).
A comparison with the VGGT pipeline~\cite{wang2025vggt} on the \realestateTenk{} dataset shows that we achieve a slightly higher \auc{30} (Ours: 88.4 vs. \vggtff{} + BA: 87.1) and an improvement of  10 AUC scores for \auc{3} (Ours: 59.1 vs. \vggtff{}: 49.1). Using their point initialization but applying our refinement (\vggtptsOurs{}) still lags behind our full approach (Ours global: \auc{30} of 88.4 vs. \vggtptsOurs{}: \auc{30} of 85.4), indicating that our point selection strategy is crucial for fine-grained performance. Interestingly, applying optimization after VGGT points initialization can worsen \auc{3} (\vggtff{}: 49.1 vs. \vggtff{} + BA: 46.7 and \vggtptsOurs{}: 46.9), highlighting its sensitivity to input quality.

\begin{table}[t]
\small
\centering
\setlength{\tabcolsep}{4pt}
\caption{
Quantitative comparison on MVS (DTU, 10 images) and RealEstate10k between our global configuration and the feed forward VGGT (VGGT ff) and VGGT + their bundle adjustment (VGGT ff + BA) according to \citee{wang2025vggt}. Finally, we report results for our pipeline initialized from VGGT points (\vggtptsOurs). Our methods are highlighted in gray. Upwards arrows represent that a higher value is better.
}
\label{tab:comparison:global_results_dtu_mvs}
\begin{tabular}{ccc}
\toprule
 &  \auct{30} $\uparrow$ & \auct{3} $\uparrow$ \\
\midrule
\multicolumn{3}{c}{MVS (DTU, 10 images per scene)} \\
\midrule
\rowMarked Ours incremental & 96.2 & 91.9  \\
\rowMarked Ours global & 99.4 & 94.1  \\
\vggtff{}~\cite{wang2025vggt} & 99.4 & 94.2   \\
\vggtff{} + BA~\cite{wang2025vggt} & 96.1 & 67.0 \\
\vggtptsOurs & 96.1 & 67.1 \\
\midrule
\multicolumn{3}{c}{RealEstate10k (124 scenes, 10 images per scene)} \\
\midrule
\rowMarked Ours incremental & 67.4 & 41.4 \\
\rowMarked Ours global & 88.4 & 59.1 \\
\vggtff{}~\cite{wang2025vggt} & 84.0 & 49.1  \\
\vggtff{} + BA~\cite{wang2025vggt}  & 87.1 & 46.7 \\
\vggtptsOurs & 85.4 & 46.9 \\
\midrule
\bottomrule
\end{tabular}
\end{table}

\subsection{\FISHEYESETTING{}}

\begin{figure}[ht]
\begin{center}
\begin{subfigure}[b]{0.95\columnwidth}
    \includegraphics[width=\linewidth]{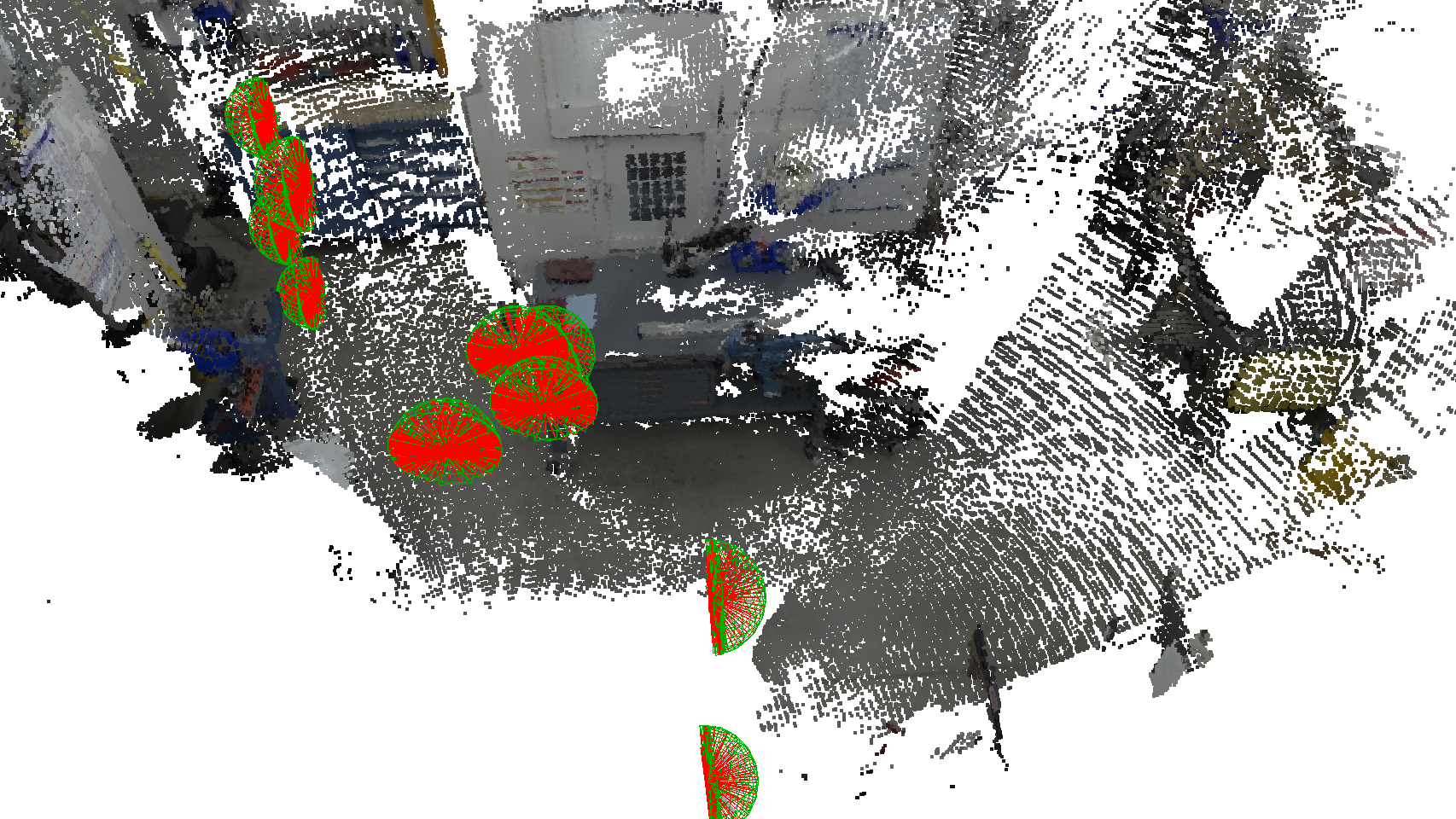}
\end{subfigure}

\begin{subfigure}[b]{0.17\columnwidth}
    \includegraphics[width=\linewidth]{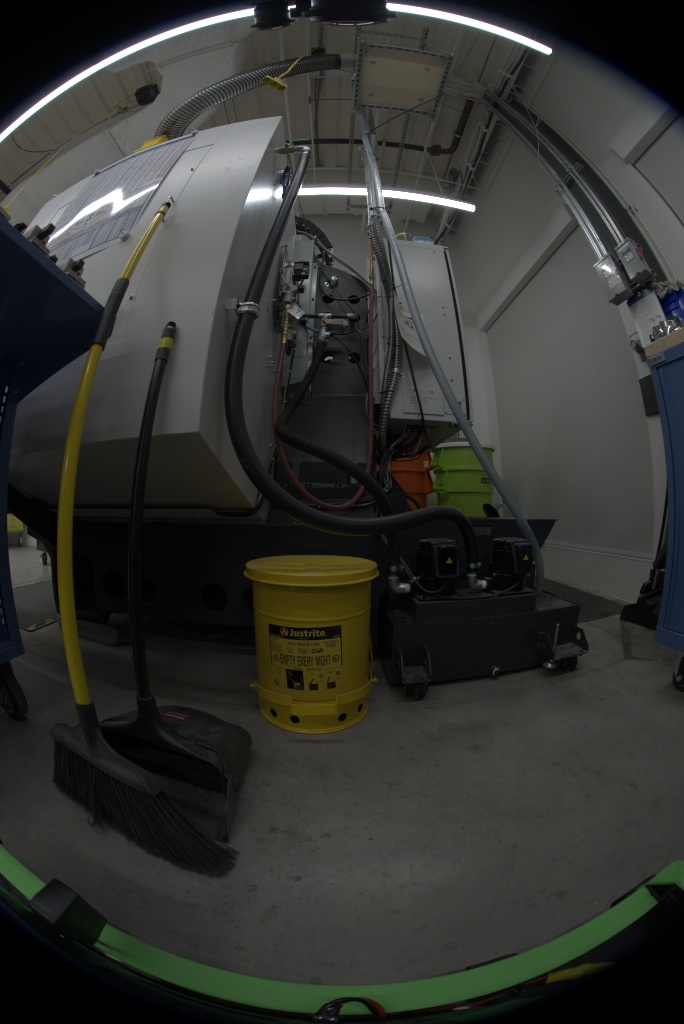}
\end{subfigure}
\begin{subfigure}[b]{0.17\columnwidth}
    \includegraphics[width=\linewidth]{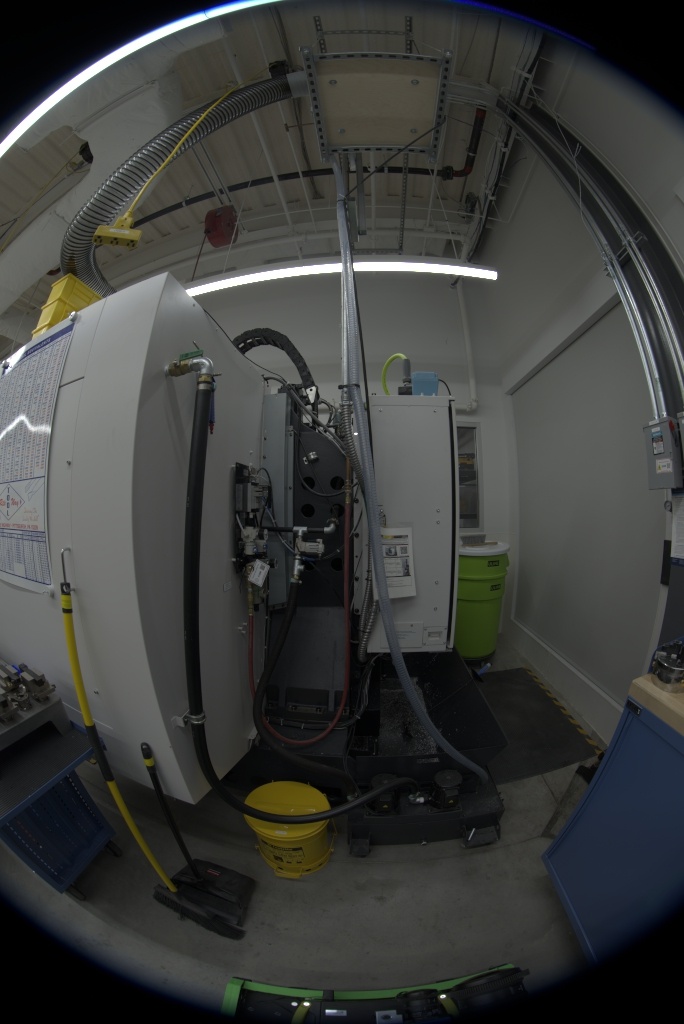}
\end{subfigure}
\begin{subfigure}[b]{0.17\columnwidth}
    \includegraphics[width=\linewidth]{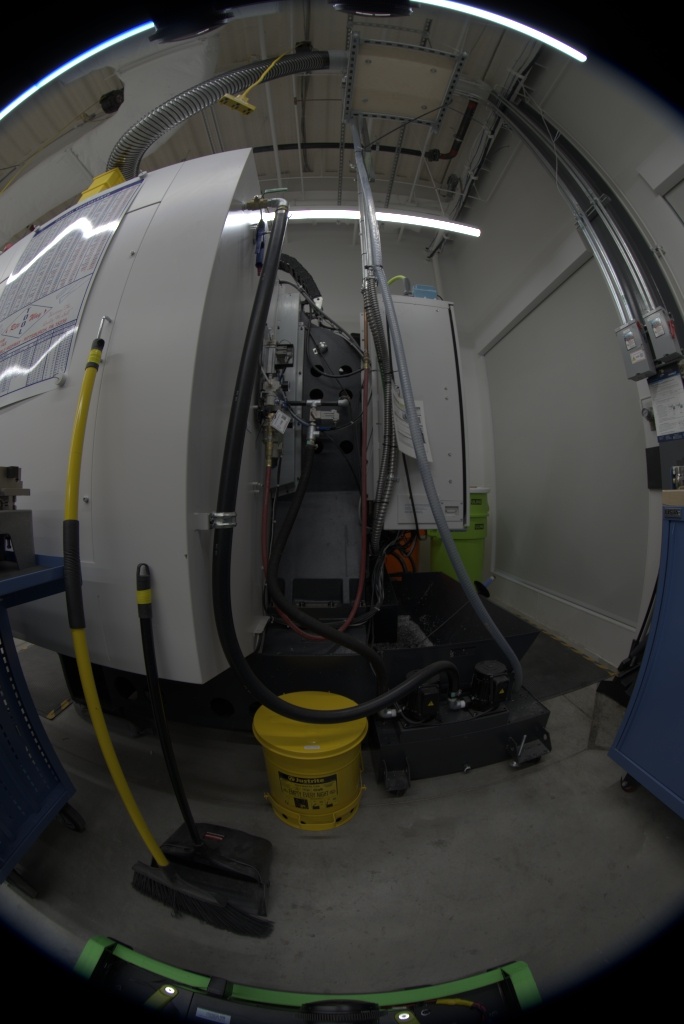}
\end{subfigure}
\begin{subfigure}[b]{0.17\columnwidth}
    \includegraphics[width=\linewidth]{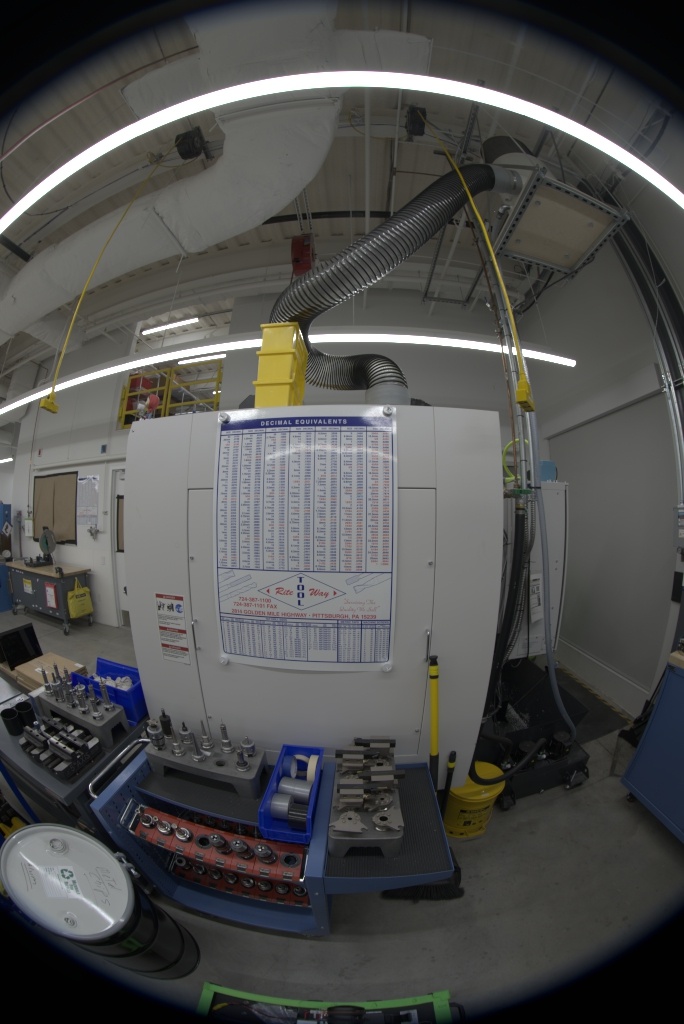}
\end{subfigure}
\begin{subfigure}[b]{0.17\columnwidth}
    \includegraphics[width=\linewidth]{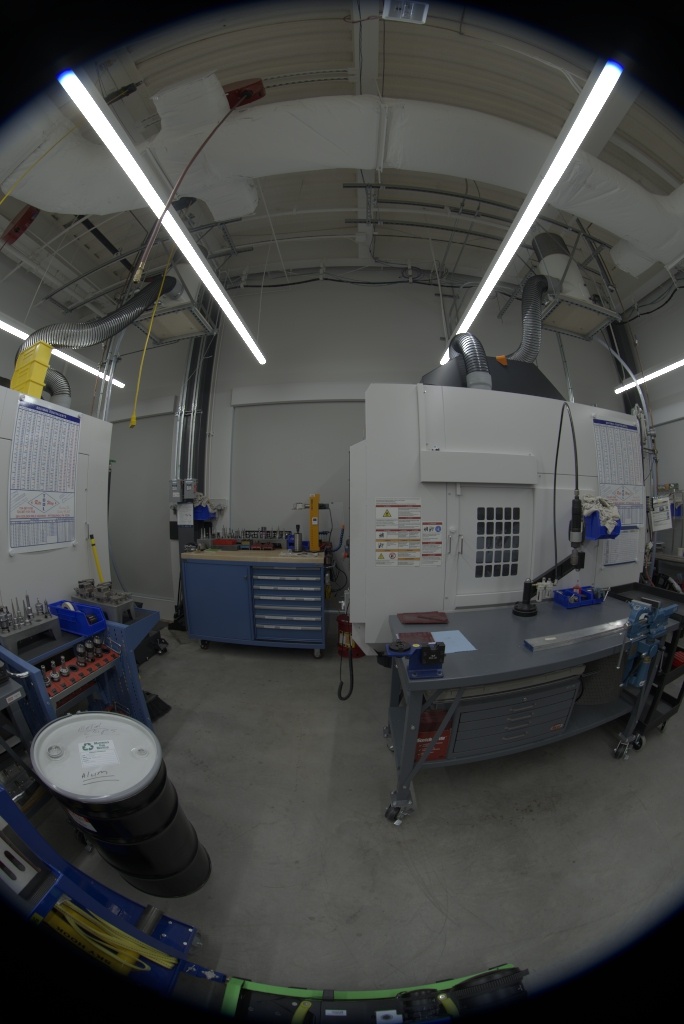}
\end{subfigure}

\begin{subfigure}[b]{0.17\columnwidth}
    \includegraphics[width=\linewidth]{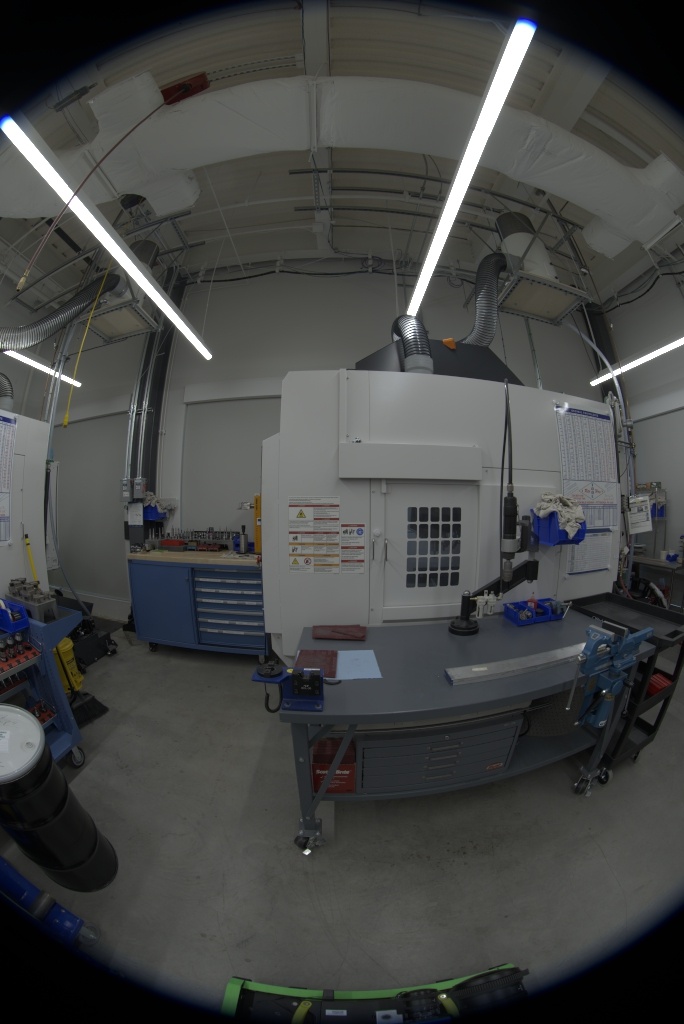}
\end{subfigure}
\begin{subfigure}[b]{0.17\columnwidth}
    \includegraphics[width=\linewidth]{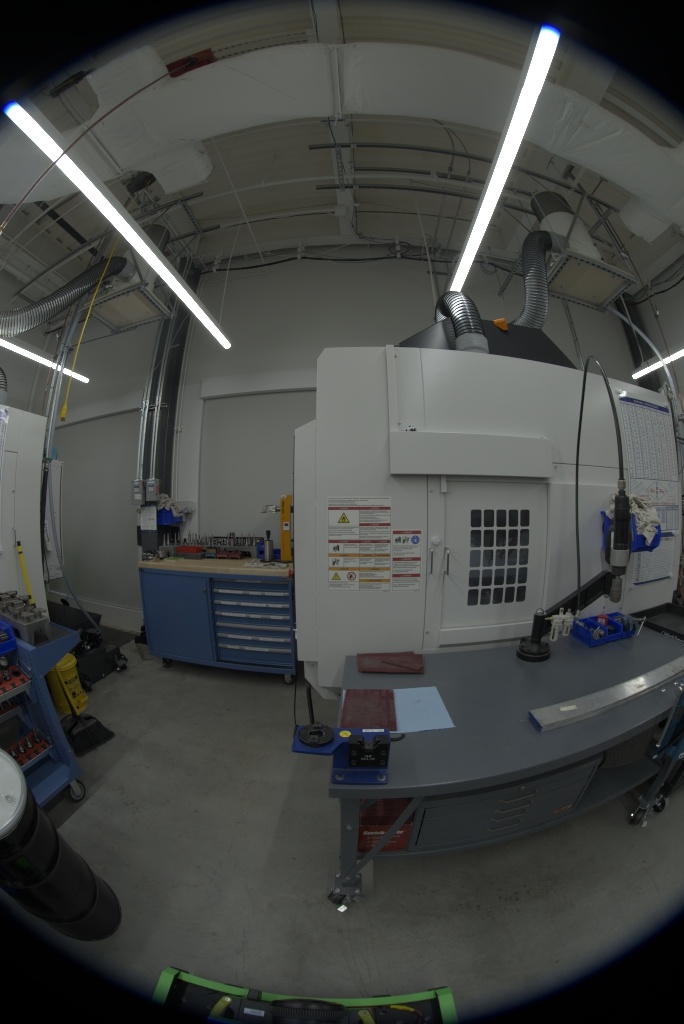}
\end{subfigure}
\begin{subfigure}[b]{0.17\columnwidth}
    \includegraphics[width=\linewidth]{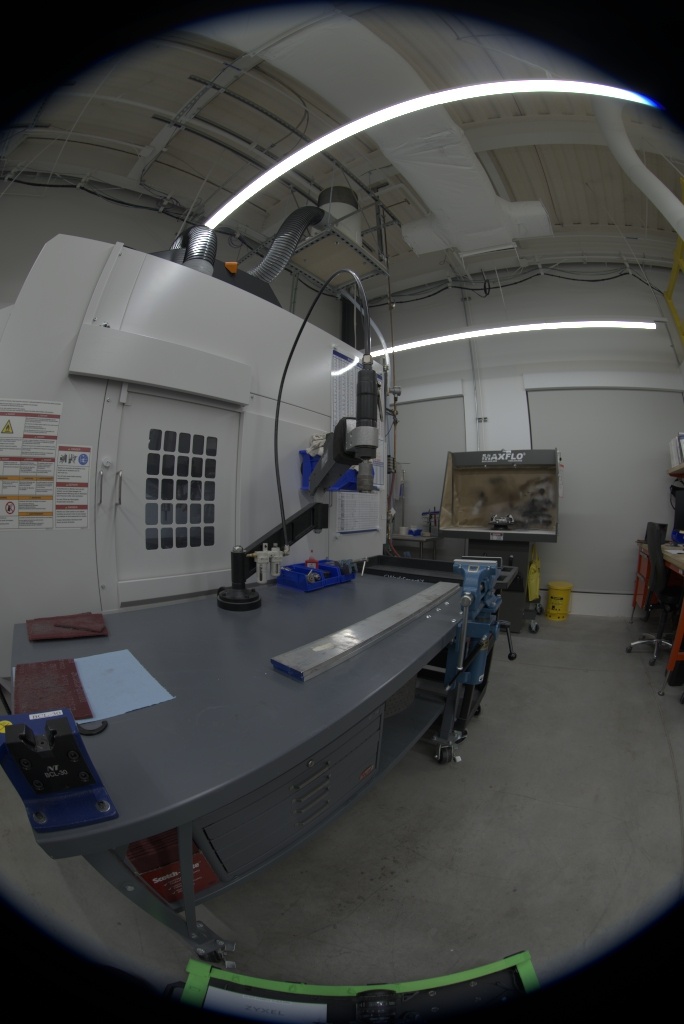}
\end{subfigure}
\begin{subfigure}[b]{0.17\columnwidth}
    \includegraphics[width=\linewidth]{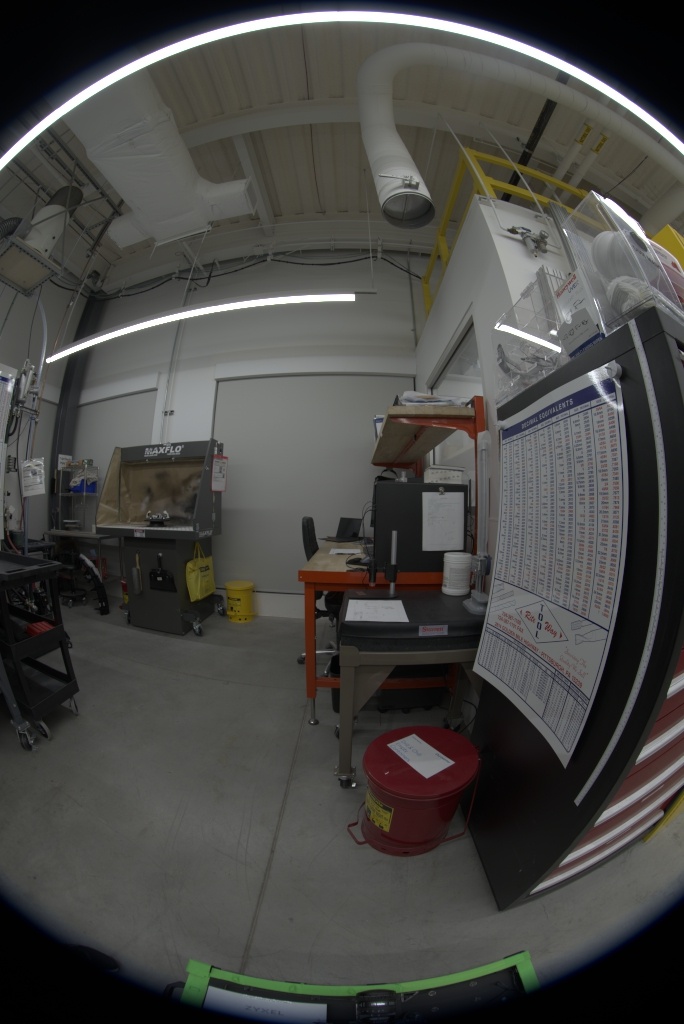}
\end{subfigure}
\begin{subfigure}[b]{0.17\columnwidth}
    \includegraphics[width=\linewidth]{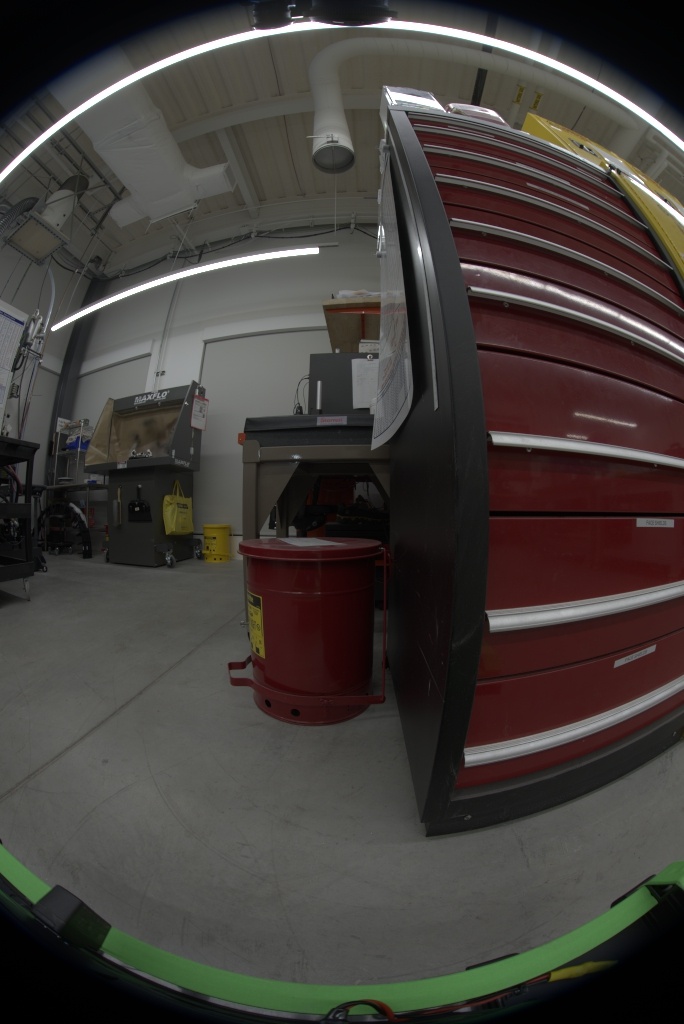}
\end{subfigure}

\caption{Example of result on the Eyeful Tower dataset \cite{VRNeRF}. Top: Triangulated 3D structure, and the 10 estimated cameras. Cameras are drawn with red rays emanating from the projection centre onto a sphere representing the normalized image plane (in green). Bottom: Input images.}
\label{fig:teaser}
\end{center}
\end{figure}
\label{sec:experiments:fisheye_settings}

\paragraph{Dataset.}
To evaluate the robustness of our method under significant difficult scenarios, e.g., severe radial distortion, we evaluate on the VR-NeRF Eyeful Tower dataset \cite{VRNeRF} - see Fig.~\ref{fig:teaser} for an example reconstruction using our method. 

The {\it VR-NeRF Eyeful Tower} dataset consists of scenes captured by a movable ``tower`` equipped with multiple cameras including pinhole and wide-angle lenses. Five scenes were captured using 9 fisheye cameras, while six scenes were captured using 22 pinhole cameras. We restrict our experiments to the fisheye subset since we explore cameras similar to their pinhole cameras in our \regularsetting{} and we want to showcase our pipeline's capability for cameras with strong radial distortion. 

From each fisheye scene, we uniformly sample 500 images using the 1K JPEG image samples (with a resolution of $684 \times 1024$ pixels). We then construct a weighted graph where each node corresponds to an image, and edge weights represent the number of 2D point correspondences between image pairs (derived from the authors' COLMAP-based ``ground truth"). To partition this graph, we employ the KaFFPa (Karlsruhe Fast Flow Partitioner) \cite{sandersschulz2013_kahip} with settings \texttt{mode=STRONG} and imbalance equal to $0.03$. This yields clusters of approximately 10 images each (with occasional clusters of size 9 or 11). For each of the five fisheye scenes, we sample 10 such clusters to be used as view sets in the evaluation.

Since our default settings in Sec.~\ref{sec:experiments:pinhole_cameras} are tuned for minimally distorted images, we test different settings for the outlier removal thresholds \Berror{} and \Ferror{}, see Sec.~\ref{sec:implementation_details} for details about the thresholds and Sec.~\ref{sec:ablation:fisheye_settings} for detailed results. 

\paragraph{Results.}
We evaluate both the incremental and global variant of our pipeline, cf. Tab. ~\ref{tab:comparison:fisheye_summary}. Our best setting for the global pipeline achieves an \auc{30} of $79.9$ and in the incremental setting we obtain an \auc{30} of $70.4$. 
Using \vggtff{}, we obtain a significantly lower \auc{30} score of 40.4, and a much lower \auc{3} of 1.2 (vs. Ours global: 61.7 and Ours incremental: 60.7). When we instead use the prediction track from VGGT, along with their point initialization from estimated cameras and depth maps (see \citee{wang2025vggt} for details), and then apply our optimization steps on top (\vggtptsOurs{}), the results improve slightly in both \auc{30} (\vggtff{}: 40.4 vs. \vggtptsOurs{}: 46.0) and \auc{3} (\vggtff{}: 1.2 vs. \vggtptsOurs{}: 1.4). However, they remain substantially worse than those achieved by our full \globalpipe{} and \incremental{} pipelines (see Section~\ref{sec:ablation:fisheye_settings} for details).

\begin{table}[t]
\centering
\small
\caption{
Incremental and global results on the \eyeful-fisheye dataset. We also compare to the feed-forward VGGT (VGGT ff) from~\citee{wang2025vggt}. Aditionally, we report results for our pipline initialized from VGGT points (\vggtptsOurs). Our methods are highlighted in gray. Upwards arrows represent that a higher value is better. 
}
\label{tab:comparison:fisheye_summary}
\begin{tabular}{ccc}
\toprule
Method & \auct{30} $\uparrow$ & \auct{3} $\uparrow$ \\
\midrule
\rowMarked Ours incremental & 70.4 & 60.7 \\
\rowMarked Ours global & 79.9 & 61.7 \\
\vggtff{} \cite{wang2025vggt} & 40.4 & 1.2 \\
\vggtptsOurs & 46.0 & 1.4 \\
\bottomrule
\end{tabular}
\end{table}

\subsection{Ablation Studies}
In this section we present the result of our ablation studies in the \regularsetting{} and in the \fisheyesetting{}. We conclude the section with a comparison across the datasets.

\subsubsection{\REGULARSETTING{}}
\label{sec:ablation_pinhole}

We conduct ablation studies on both the \globalpipe{} and \incremental{} pipelines using the \mvs{} dataset from \citee{jensen2014large} and the \realestateTenk{} dataset from \citee{zhou2018stereo_real_estate_10k}.

For both datasets, \mvs{} and \realestateTenk{}, we evaluate the Cartesian product of the following parameter settings in the incremental and global reconstruction pipelines:
\begin{itemize}
    \item \topk{} $\in \{0,1,2,3\}$, 
    \item \gridsize{} $\in \{20,40,80\}$, 
    \item \probscore{} $\in \{\text{True}, \text{False}\}$.
\end{itemize}
In the incremental pipeline we also evaluate if the \bruteforce{} strategy improves the results. 

\paragraph{\realestateTenk{}.}

Table~\ref{tab:realestate_incremental} shows the top configuration, followed by variations to individual components (the underlined ones in the tables) in the incremental pipeline. The largest drop in \auc{30} occurs when disabling angle-based sampling ($\topk = 0$), decreasing performance by 20.1 points. Disabling the \bruteforce{} strategy also significantly reduces \auc{30} (-12.6). Lowering the \gridsize{} from 80 to 40 or 20 impacts both \auc{30} and \auc{3}, while refining the principal point results in the largest decrease in \auc{3} (-13.7).

\begin{table}[t]
\small
\caption{Ablation study on the \realestateTenk{} dataset using the \incremental{} pipeline (averaged over 5 runs).
The top row reports the best-performing configuration. Subsequent rows show the results of varying one parameter (the underlined one) at a time to isolate its effect. Abbreviations: BF: \bruteforce{}, CS: \gridsize{}, PS: \probscore{}, k: \topk, RP: refine principal point, T: True, F: False.}
\label{tab:realestate_incremental} \centering
\begin{tabular}{|c|c|c|c|c|c|c|}
\hline
\multicolumn{7}{|c|}{\Incremental{} Pipeline, \realestateTenk{}} \\
\hline
BF & CS & k & PS & RP & \auct{30} & \auct{3} \\
\hline
\rowMarked T & 80 & 3 & T & F & 67.4 & 41.4 \\
T & 80 & \underline{2} & T & F & 67.0 & 41.2 \\
T & 80 & \underline{1} & T & F & 66.7 & 40.9 \\
T & 80 & 3 & \underline{F} & F & 66.3 & 40.2 \\
T & \underline{40} & 3 & T & F & 65.2 & 40.1 \\
T & 80 & 3 & T & \underline{T} & 64.7 & 27.7 \\
T & \underline{20} & 3 & T & F & 64.3 & 39.7 \\
\underline{F} & 80 & 3 & T & F & 54.8 & 33.4 \\
T & 80 & \underline{0} & T & F & 47.3 & 29.8 \\
\hline
\end{tabular}
\end{table}
\begin{table}[t]
\caption{
Ablation study on \realestateTenk{} dataset using the \globalpipe{} pipeline (averaged over 5 runs).
The top row reports the best-performing configuration. Subsequent rows show the results of varying one parameter (the underlined one) at a time to isolate its effect. Abbreviations: CS: \gridsize{}, k: \topk, PS: \probscore{}, RF, RD, RP: refine focal, distortion and principal point, respectively.
}
\label{tab:realestate_global} 
\centering
\small
\begin{tabular}{|c|c|c|c|c|c|}
\hline
\multicolumn{6}{|c|}{\Globalpipe{} Pipeline, \realestateTenk{}} \\
\hline
CS & k & PS & step & \auc{30} & \auc{3} \\
\hline
\rowMarked 40 & 1 & T & RF & 88.4 & 59.1 \\
\underline{80} & 1 & T & RF & 88.3 & 58.7 \\
40 & \underline{0} & T & RF & 88.2 & 59.4 \\
40 & 1 & \underline{F} & RF & 88.1 & 58.8 \\
40 & \underline{3} & T & RF & 88.0 & 59.0 \\
40 & \underline{2} & T & RF & 88.0 & 58.9 \\
40 & 1 & T & \underline{RD} & 87.9 & 54.2 \\
\underline{20} & 1 & T & RF & 87.8 & 59.5 \\
40 & 1 & T & \underline{RP} & 84.5 & 34.5 \\
\hline
\end{tabular}
\end{table}

For the global pipeline (Table~\ref{tab:realestate_global}), \topk{} and \gridsize{} have smaller effects. However, principal point refinement (column RP) still notably reduces \auc{3} (-24.6), suggesting a sensitivity to fine-scale accuracy.

\paragraph{\mvs{}.}

Table~\ref{tab:mvs_incremental} shows the incremental pipeline results on \mvs{}. Reducing \gridsize{} from 80 to 20 leads to a 9.2-point drop in \auc{30}, while removing \bruteforce{} lowers it by 4.6 points. Interestingly, disabling the angular sampling ( $\topk = 0$) has limited impact here.

\begin{figure}[t]
    \centering

    \begin{subfigure}[b]{0.3\columnwidth}
        \includegraphics[width=\linewidth]{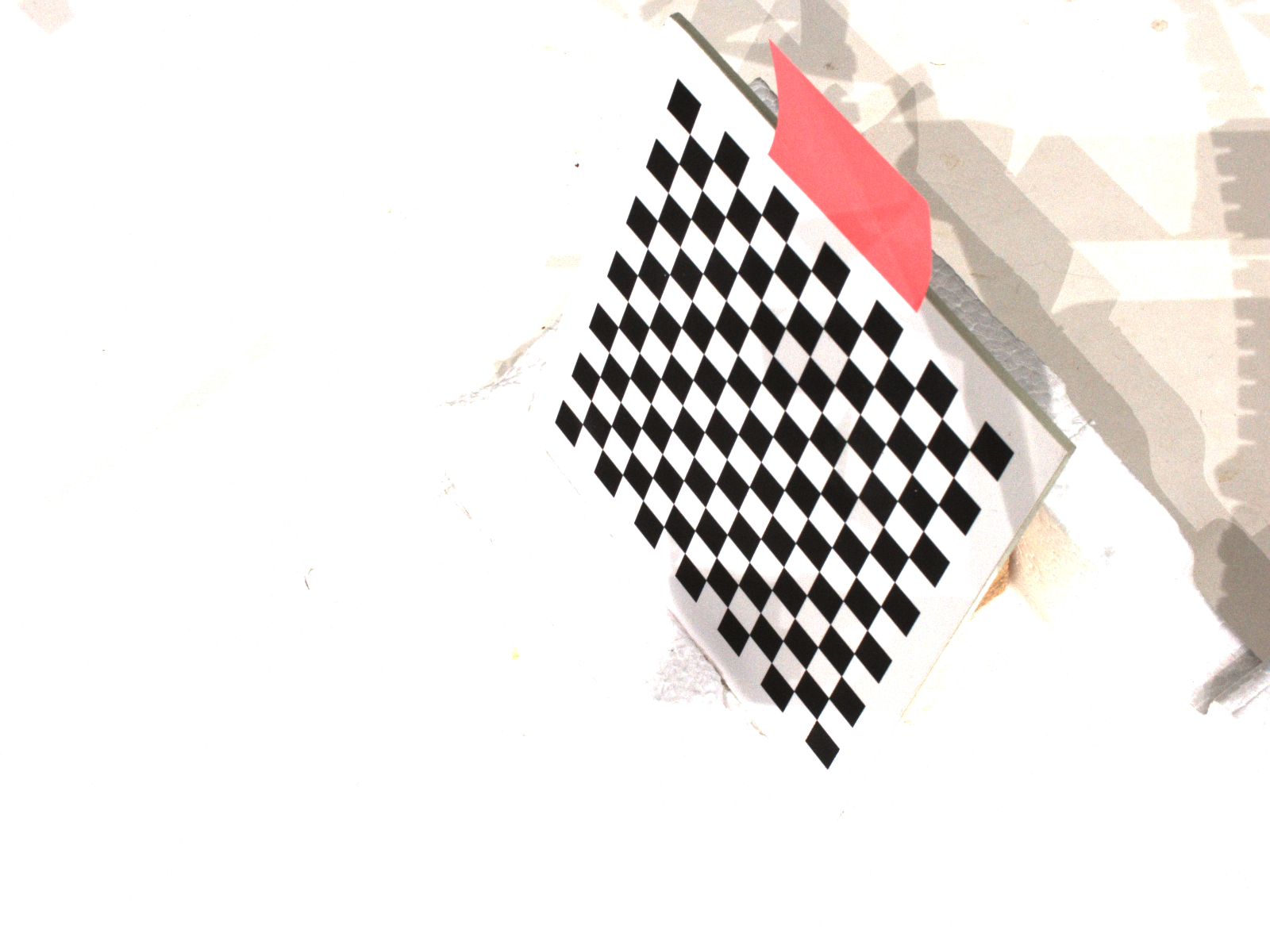}
    \end{subfigure}
    \hfill
    \begin{subfigure}[b]{0.3\columnwidth}
        \includegraphics[width=\linewidth]{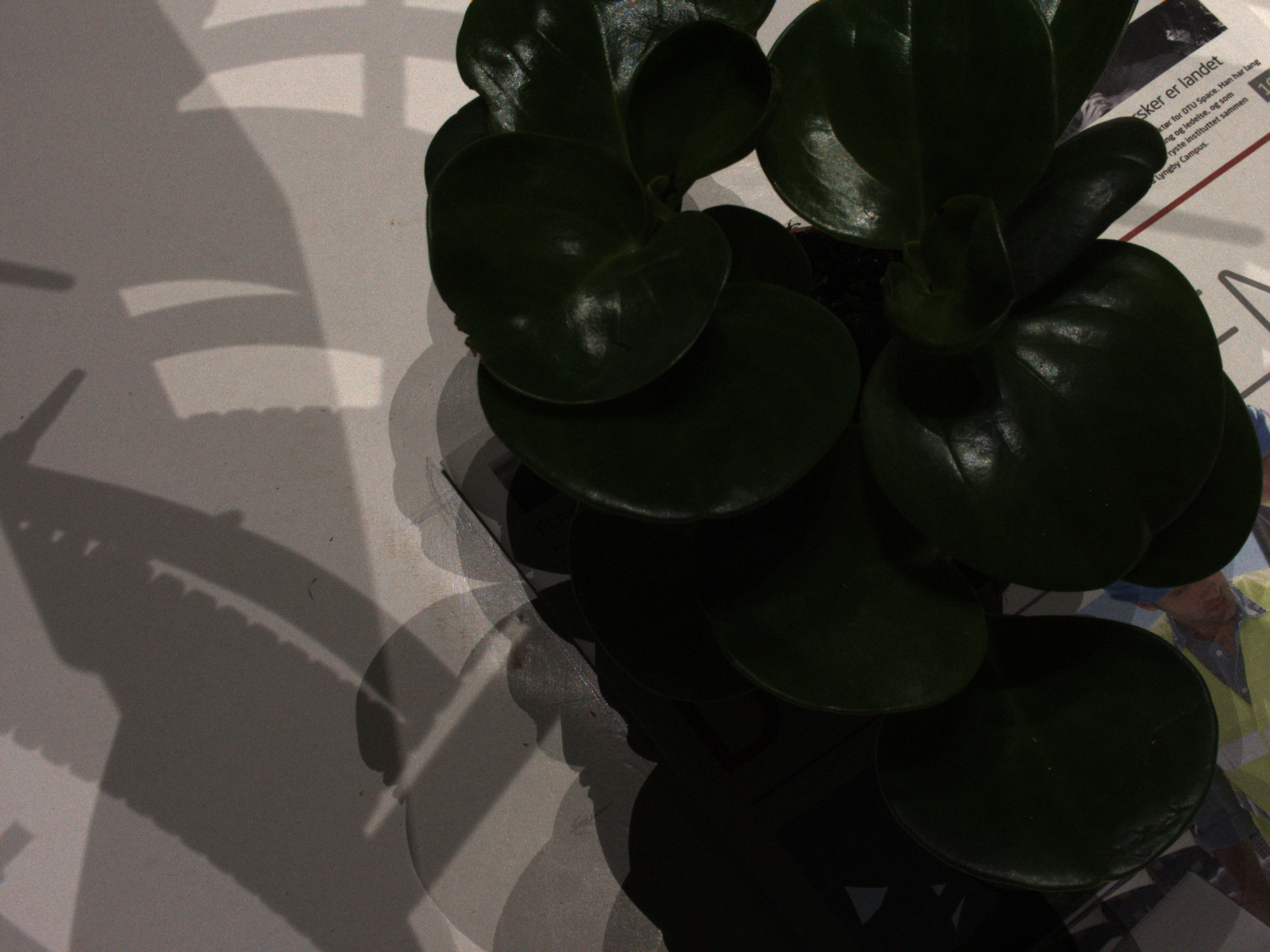}
    \end{subfigure}
    \hfill
    \begin{subfigure}[b]{0.3\columnwidth}
        \includegraphics[width=\linewidth]{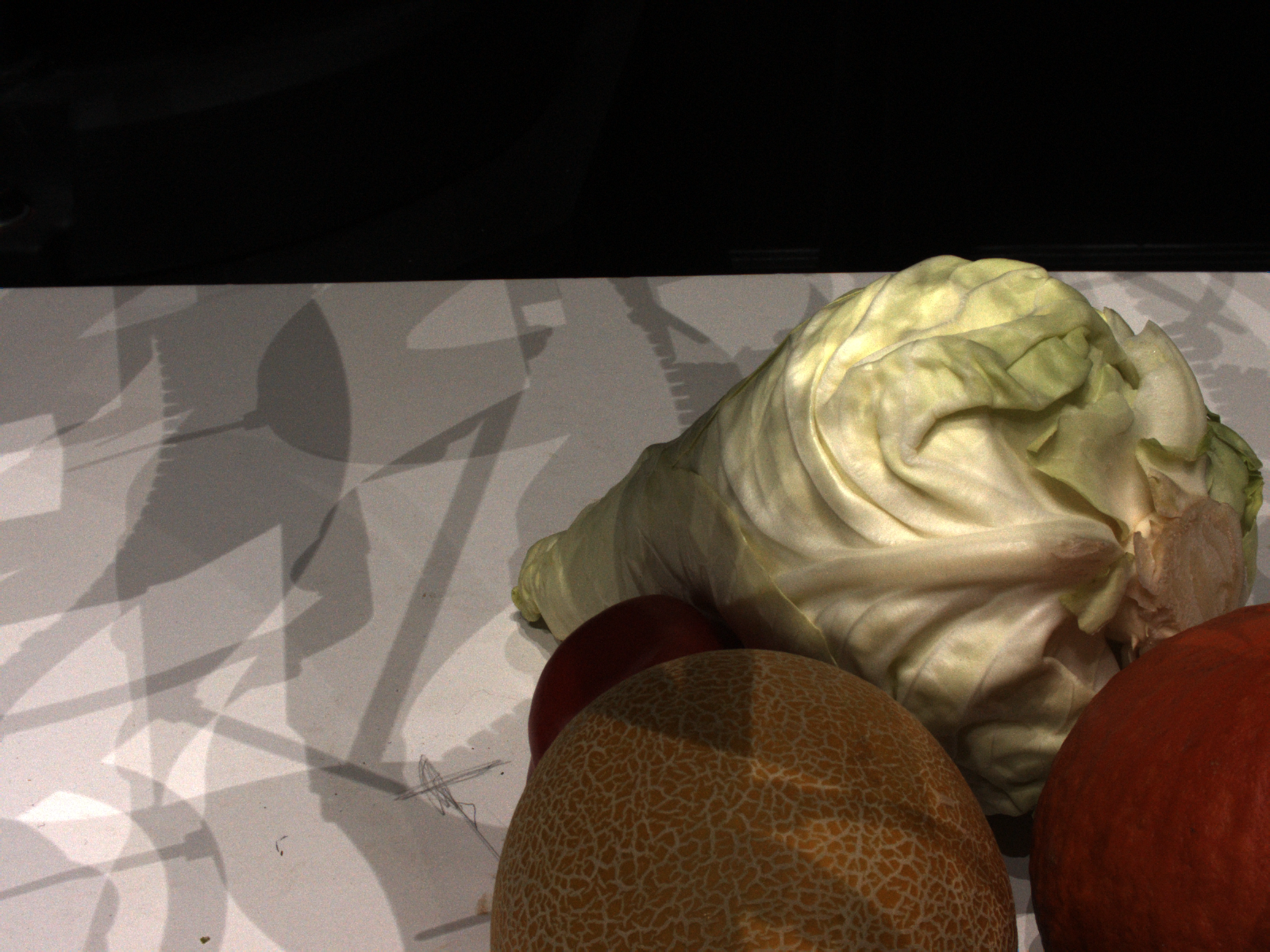}
    \end{subfigure}

    \vspace{0.5em}

    \begin{subfigure}[b]{0.3\columnwidth}
        \includegraphics[width=\linewidth]{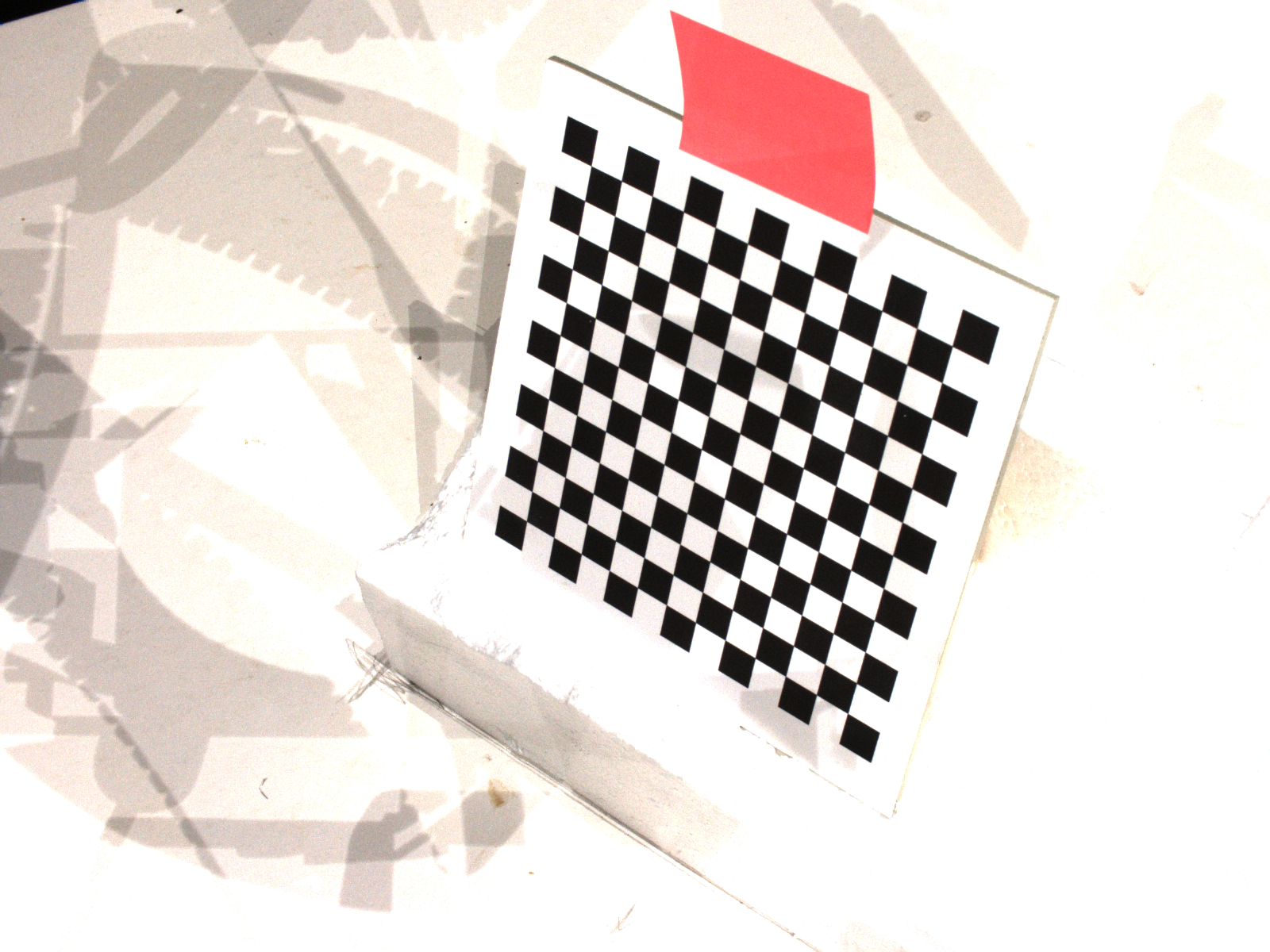}
        \caption{}
        \label{fig:mvs-failure-cases:scene_54}
    \end{subfigure}
    \hfill
    \begin{subfigure}[b]{0.3\columnwidth}
        \includegraphics[width=\linewidth]{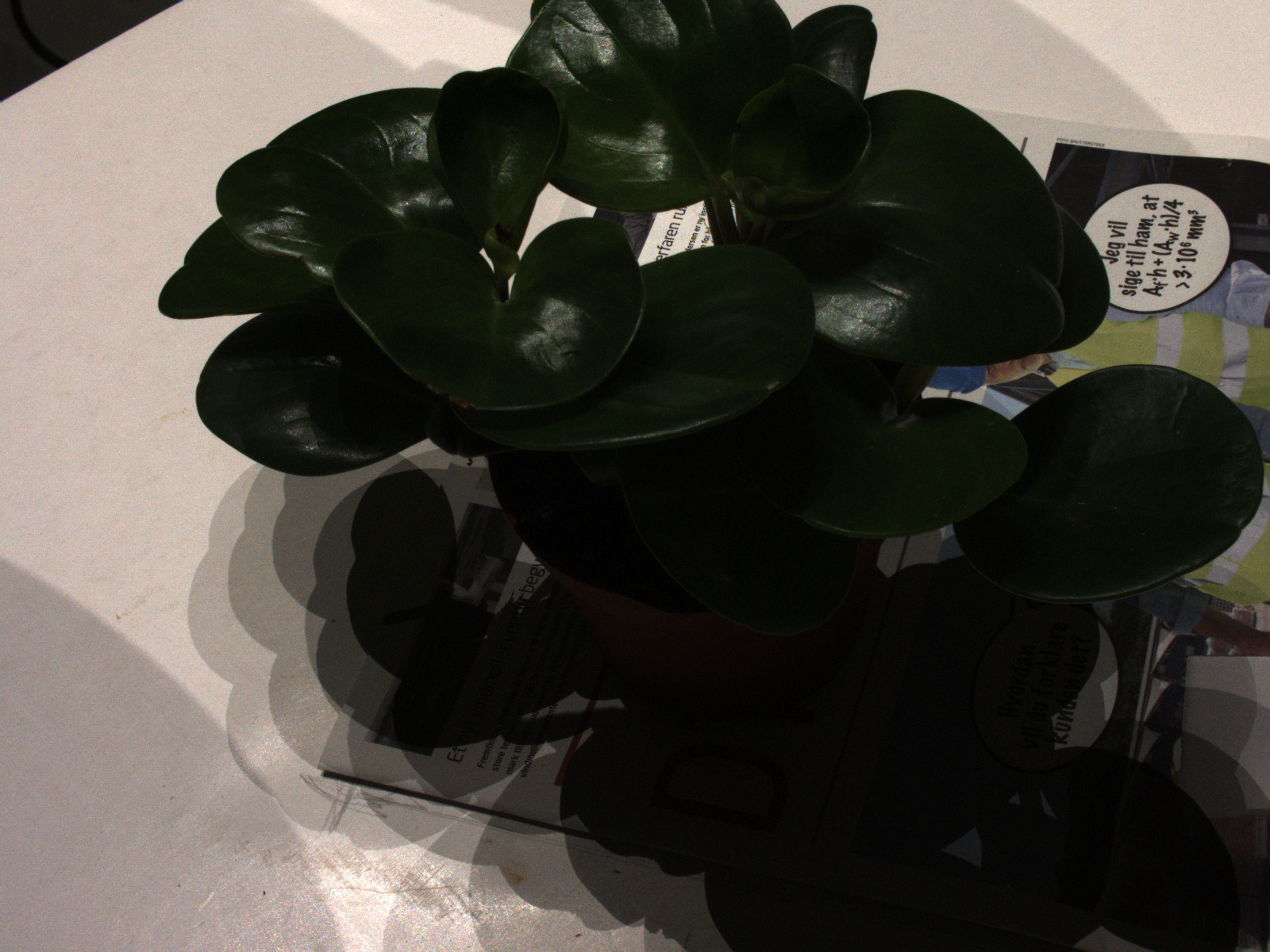}
        \caption{}
    \end{subfigure}
    \hfill
    \begin{subfigure}[b]{0.3\columnwidth}
        \includegraphics[width=\linewidth]{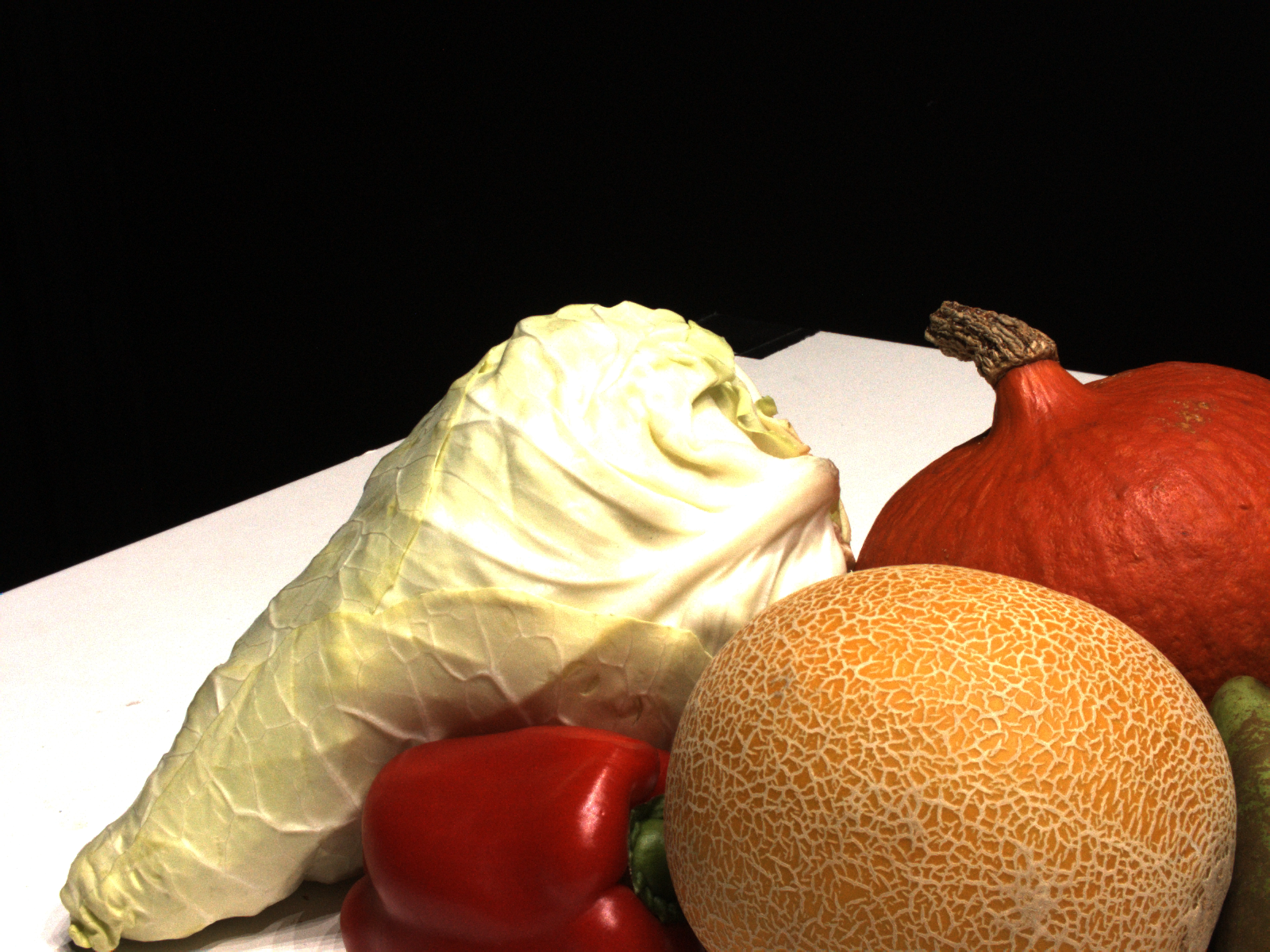}
        \caption{}
    \end{subfigure}

    \caption{Example images from failure cases on the \mvs{} dataset \cite{jensen2014large} in the \incremental{} pipeline. These scenes contain shadows that vary from frame to frame and a completely white table surface, which likely contribute to the observed reconstruction failures. In (a), the flat checkerboard also constitutes a dominant scene plane -- a known challenging case in SFM.}
    \label{fig:mvs-failure-cases}
\end{figure}

Among the 124 scenes in the dataset, seven exhibit \auc{30} scores below 0.1, all of which also fall below 0.6--indicating consistent failure. Many scenes include strong shadows, and notably, all failure cases share the presence of a white table surface and a fully black background. This suggests that the combination of low texture, varying shadows, and limited geometric structure may lead to degraded performance. A clear example is scene 54 (see Fig.~\ref{fig:mvs-failure-cases:scene_54}), which features a flat checkerboard placed on a white table. Due to the limited 3D structure and significant variation in shadows cast by the robot arm across views, the matching becomes unreliable. Figure~\ref{fig:mvs-failure-cases} shows three representative failure cases. Note that in our target scenario, cameras that capture scene at same time from different views, the shadow artifact will not be present. While one of the seven scenes (scene 12) also fails in the global pipeline, the others do not, suggesting that a good global initialization can mitigate these issues.

\begin{table}[t]
\small
\caption{Ablation study on the \mvs{} dataset using our \Incremental{} pipeline (averaged over 5 runs).
The top row reports the best-performing configuration. Subsequent rows show the results of varying one parameter (the underlined one) at a time to isolate its effect. Abbreviations: BF: \bruteforce{}, CS: \gridsize{}, k: \topk, PS: \probscore{}, RP: refine principal point, T: True, F: False.}
\label{tab:mvs_incremental} 
\centering
\begin{tabular}{|c|c|c|c|c|c|c|}
\hline
\multicolumn{7}{|c|}{\Incremental{} Pipeline, \mvs{}} \\
\hline
BF & CS & k & PS & RP  & \auct{30} & \auct{3} \\
\hline
\rowMarked T & 80 & 3 & F & F & 96.2 & 91.9 \\
T & 80 & \underline{0} & F & F & 95.6 & 91.3 \\
T & 80 & \underline{2} & F & F & 95.2 & 90.5 \\
T & 80 & \underline{1} & F & F & 94.6 & 90.6 \\
T & 80 & 3 & \underline{T} & F & 94.5 & 90.5 \\
T & \underline{40} & 3 & F & F & 93.8 & 89.9 \\
\underline{F} & 80 & 3 & F & F & 91.6 & 87.3 \\
T & 80 & 3 & F & \underline{T} & 90.6 & 54.1 \\
T & \underline{20} & 3 & F & F & 87.0 & 83.9 \\
\hline
\end{tabular}
\end{table}

In comparison, Table~\ref{tab:mvs_global}, which reports ablations using our \globalpipe{} pipeline, shows that the effect of varying the \gridsize{} is relatively small, but still important (e.g., down 2.5 points on \auc{30} when going from \gridsize{} 80 to 20). Refining the principal point has an impact on \auc{30} (-6.2 points), but more importantly leads to a substantial decrease in \auc{3} (-39.2 points).

\begin{table}[t]
\small
\caption{
Ablation study on the \textit{\mvs{}} dataset using our \textit{\globalpipe{}} pipeline (averaged over 5 runs).
The top row reports the best-performing configuration. Subsequent rows show the results of varying one parameter (the underlined one) at a time to isolate its effect. Abbreviations: CS: \gridsize{}, k: \topk, PS: \probscore{}, RF, RD, RP: refine focal, distortion and principal point, respectively. 
}
\label{tab:mvs_global} \centering
\begin{tabular}{|c|c|c|c|c|c|}
\hline
\multicolumn{6}{|c|}{\Globalpipe{} Pipeline, \mvs{}} \\
\hline
CS & k & PS & step & \auc{30} & \auc{3} \\
\hline
\rowMarked 80 & 3 & T & RE & 99.4 & 94.1 \\
80 & \underline{1} & T & RE & 99.3 & 94.0 \\
80 & \underline{2} & T & RE & 99.3 & 94.0 \\
80 & 3 & T & \underline{RD} & 99.3 & 94.6 \\
80 & 3 & T & \underline{RF} & 99.2 & 92.8 \\
80 & \underline{0} & T & RE & 99.2 & 93.9 \\
80 & 3 & \underline{F} & RE & 99.2 & 93.8 \\
\underline{40} & 3 & T & RE & 98.7 & 93.5 \\
\underline{20} & 3 & T & RE & 96.9 & 91.9 \\
80 & 3 & T & \underline{RP} & 93.2 & 54.9 \\
\hline
\hline
\end{tabular}
\end{table}

\subsubsection{\FISHEYESETTING{}}
\label{sec:ablation:fisheye_settings}

\begin{table}[t]
\small
\caption{Ablation study using our \incremental{} pipeline on the \eyeful{} dataset (averaged over 5 runs). Top row shows the best result; other rows show the effect of changing one parameter (the underlined one). Abbreviations: BF: \bruteforce{}, CS: \gridsize{},  RP: refine principal point, NRP: don't refine principal point, RCT: refine camera type. The steps are in the following order in the pipeline: NRP, RP, RCT; see Sec.~\ref{sec:implementation_details} for details.}
\label{tab:ablation_fisheye_incremental} \centering
\begin{tabular}{|c|c|c|c|c|c|c|}
\hline
\multicolumn{7}{|c|}{\Incremental{} Pipeline, \eyeful{}} \\
\hline
BF & CS  & step & $\mathcal{B}_\mathrm{err}$ & $\mathcal{F}_\mathrm{err}$ & \auct{30} & \auct{3} \\
\hline
\rowMarked T & 80  & RCT & 100 & $\infty$ & 70.4 & 60.7 \\
T & 80  & \underline{RP} & 100 & $\infty$ & 70.4 & 60.7 \\
T & 80  & \underline{NRP} & 100 & $\infty$ & 69.9 & 52.8 \\
T & 80  & RCT & 100 & \underline{250} & 68.9 & 58.1 \\
T & 80  & RCT & 100 & \underline{500} & 67.9 & 57.4 \\
T & 80  & RCT & \underline{20} & $\infty$ & 67.7 & 56.5 \\
T & 80  & RCT & \underline{$\infty$} & $\infty$ & 67.3 & 50.5 \\
T & 80  & RCT & \underline{10} & $\infty$ & 66.7 & 56.4 \\
\underline{F} & 80  & RCT & 100 & $\infty$ & 66.5 & 55.2 \\
T & 80  & RCT & 100 & \underline{100} & 65.7 & 53.7 \\
T & \underline{40}  & RCT & 100 & $\infty$ & 62.6 & 52.5 \\
T & 80  & RCT & 100 & \underline{10} & 53.9 & 39.2 \\
T & \underline{20}  & RCT & 100 & $\infty$ & 52.1 & 44.5 \\
\hline
\end{tabular}
\end{table}

Table~\ref{tab:ablation_fisheye_incremental} shows the top-performing configuration (top row), followed by an ablation study where we vary one component at a time (the underlined ones in the table) in the incremental pipeline on the \eyeful{} dataset. The best performance is achieved when \Ferror{} is set to $\infty$, which effectively disables outlier removal based on the initial fundamental matrix estimation. This makes sense because corresponding points in undistorted fisheye images can strongly violate the fundamental matrix constraint. 
Moreover, thanks to the robust cost function used in our bundle adjustment, retaining a small number of incorrect correspondences is less detrimental, while the preceding cycle-based filtering step (see~\ref{sec:correspondance_cycle_sampling}) increases the proportion of reliable correspondences.

Reducing \Ferror{} to 10 (as done in the \regularsetting{}) has a severe negative impact on both \auc{30} and \auc{3}, decreasing them by 16.5 and 21.5 points, respectively. A reduction to 100 also degrades performance significantly, although less so than a reduction to 10. In contrast, reducing \Ferror{} to 250 or 500 results in smaller performance drops.

Lowering the \gridsize{} to 20 leads to a notable decline in both \auc{30} and \auc{3}, though the effect is slightly less severe at \gridsize{} = 40. Similarly, setting \Berror{} too low negatively affects both AUC metrics, although the impact is less critical than with \Ferror{}.

It is also worth noting that the best results are obtained using the \textit{Refine Camera Type (RCT)} setting. Interestingly, the same performance is achieved in the preceding step where only the principal point is refined (RP).

For our global pipeline (see Table~\ref{tab:ablation_fisheye_global}), we observe similar trends. The best configuration again involves disabling outlier removal from the initial fundamental matrix estimation.

In this setting, a \gridsize{} of 40 yields the best result. Using a grid size of 80 gives slightly lower performance, while reducing it to 20 causes a 4.5-point drop in \auc{30}. As with the incremental pipeline, skipping refinement of the camera center leads to a substantial decrease in \auc{30}.

Table~\ref{tab:comparasion_fisheye_vggt} presents results using only VGGT ff and our combined method that incorporates point optimization on top of VGGT (\vggtptsOurs{}). During our experiments with our \globalpipe{} pipeline, we found that having a high \Ferror{} improved the camera pose estimation. To study this, we adjust the reprojection error threshold (denoted as \rmaxx{} in the table). In \citee{wang2025vggt}, 3D points and their tracks with keypoint reprojection errors greater than 12 pixels are removed prior to bundle adjustment (see also \citee{facebookresearch_vggt_evaluation_2025} for details). We evaluate both this original setting and an alternative where outlier removal is entirely disabled. While removing this filtering step slightly improves performance (\rmaxx{} = 12: \auc{30} of 43.5 vs. \rmaxx{} = $\infty$: \auc{30} of 46.0), it still lags significantly behind our method. This is especially evident in the \auc{3} metric, where the best setting for \vggtptsOurs{} achieves only 1.4, whereas our \globalpipe{} pipeline that uses our points reaches 61.7 (compare Tab.~\ref{tab:comparison:fisheye_summary}).

\begin{table}[t]
\small
\caption{ Ablation study on the \eyeful{} dataset using our \globalpipe{} pipeline (averaged over 5 runs). The top row reports the best-performing configuration. Subsequent rows show the results of varying one parameter (the underlined one) at a time to isolate its effect.  Abbreviations: CS: \gridsize, RCT: refine camera type, RP refine principal point. NRP: don't refine principal point. The steps are in the following order in the pipeline: NRP, RP, RCT; see Sec.~\ref{sec:implementation_details} for details.
}
\label{tab:ablation_fisheye_global}
\centering
\begin{tabular}{|c|c|c|c|c|c|}
\hline
\multicolumn{6}{|c|}{\Globalpipe{} Pipeline, \eyeful{}} \\
\hline
CS & step & $\mathcal{B}_\mathrm{err}$ & $\mathcal{F}_\mathrm{err}$ & \auct{30} & \auct{3} \\
\hline
\rowMarked 40 &  RCT & $\infty$ & $\infty$ & 79.9 & 61.7 \\
40 & \underline{RP} & $\infty$ & $\infty$ & 79.9 & 61.7 \\
\underline{80} & RCT & $\infty$ & $\infty$ & 79.6 & 58.0 \\
40 & RCT & $\infty$ & \underline{500} & 78.9 & 60.9 \\
40 & \underline{NRP} & $\infty$ & $\infty$ & 78.8 & 52.2 \\
40 & RCT & $\infty$ & \underline{10} & 78.5 & 60.7 \\
40 & RCT & $\infty$ & \underline{250} & 78.5 & 61.5 \\
40 & RCT & $\infty$ & \underline{100} & 78.3 & 60.3 \\
\underline{20} & RCT & $\infty$ & $\infty$ & 75.4 & 60.1 \\
\hline
\end{tabular}
\end{table}

\begin{table}[t]
\small
\caption{Result on \eyeful{} using \vggtff{} and \vggtptsOurs{} for diffrent settings of the reprojection error threshold \rmaxx{}.}
\label{tab:comparasion_fisheye_vggt}
\centering
\begin{tabular}{|c|c|c|c|c|c|}
\hline
\multicolumn{6}{|c|}{\vggtff{ and \vggtptsOurs{}}} \\

\hline
\multicolumn{3}{|c|}{Method} & \rmaxx{} & \auct{30} & \auct{3} \\
\hline
\multicolumn{3}{|c|}{VGGT ff} & -- & 40.4 & 1.2 \\
\multicolumn{3}{|c|}{\vggtptsOurs} & $\infty$ & 46.0 & 1.4 \\
\multicolumn{3}{|c|}{\vggtptsOurs} & $12$ & 43.5 & 0.7 \\
\hline
\end{tabular}
\end{table}

\subsubsection{Summary and Comparison Across Datasets}

Interestingly, using a \gridsize{} of 80 proved beneficial across all three datasets. A \gridsize{} of 40 also consistently outperformed a value of 20 (see Tables~\ref{tab:comparison:global_results_dtu_mvs}, \ref{tab:realestate_incremental}, and \ref{tab:ablation_fisheye_incremental}). The use of the \bruteforce{} strategy (available only in the \incremental{} pipeline) was also found to be important for all datasets.

In the \regularsetting{}, where different values of \topk{} were tested, \topk{} = 3 yielded the best results. However, values of 1 and 2 performed similarly well on both datasets. Notably, setting \topk{} = 0 caused a significant drop in performance on the \realestateTenk{} dataset, a pattern not observed on the \mvs{} dataset. A closer inspection of the \realestateTenk{} scenes suggests that many consist primarily of rotational camera motion--using \topk{}  $> 0$ helps enforce a view ordering that improves triangulation angles and, consequently, the stability of the \incremental{} pipeline.

Refining the principal point was generally detrimental--particularly for \auc{3}--on both the \realestateTenk{} and \mvs{} datasets. This may be due to insufficient calibration quality, or the lack of shared intrinsics across views, both of which are important when refining the camera center. Interestingly, however, principal point refinement improved results in both the \incremental{} and \globalpipe{} pipelines on the \eyeful{} dataset.

In the ablation study on \realestateTenk{}, a high \Ferror{} value was shown to be important for both pipelines. Similarly, a relatively high \Berror{} was beneficial in the \incremental{} pipeline. Whether it makes sense to restrict these thresholds during later refinement iterations, or instead apply stricter thresholds during intermediate steps, remains an open question. This is especially relevant considering that our point selection strategy--based on dense matching--already likely reduces the number of outliers. How best to tune these thresholds in this context is an interesting direction for future work.

\section{\uppercase{Conclusion}}
This work presents a modular and interpretable framework for multi-view camera self-calibration using dense correspondences. The framework is aimed at real-world deployments in which camera rigs are uncalibrated and individual cameras may exhibit significant distortion.
We show that robust correspondence sampling—via hierarchical cycle tests and a triangulation-based scoring scheme—can significantly enhance both incremental and global SfM pipelines. When initialized with VGGT, our method further improves accuracy across diverse datasets, including scenes with severe radial distortion.

Notably, our approach requires no additional training and still rivals or surpasses state-of-the-art feedforward models in several benchmarks. The ability to handle challenging settings like fisheye cameras without relying on heavy learned priors makes our method especially suited for scalable, field-ready applications in behavioural science and forensics. We believe our framework represents a step toward more robust and interpretable alternatives to black-box SfM solutions, and opens up new directions for hybrid pipelines that combine dense matching, heuristics and geometry.

\section*{\uppercase{Acknowledgements}}
This work was partially supported by the Wallenberg AI, Autonomous Systems and Software Program (WASP) funded by the Knut and Alice Wallenberg Foundation, Sweden, and by the Swedish Research Council, grant no.\ 2022-04206.

The computations were enabled by the Berzelius resource provided by 
the Knut and Alice Wallenberg Foundation at the National Supercomputer Centre.

To improve the grammar, clarity and flow of the text, \mbox{ChatGPT}~\cite{openai_gpt5_model_2025} was used.
The authors verified all AI-suggested edits and remain fully responsible for the content.

Special thanks to Henry Fr\"ocklin, who took the photos in Figure~\ref{fig:crime_scene}.

\bibliographystyle{apalike}
{\small\bibliography{example}}

\end{document}